\documentclass{article}

\usepackage[citestyle=authoryear-comp, sorting=nyt,maxbibnames=99,backend=biber]{biblatex}
\title{Bibliography management: \texttt{biblatex} package}

\usepackage{PRIMEarxiv}

\usepackage[utf8]{inputenc} 
\usepackage[T1]{fontenc}    
\usepackage{hyperref}       
\usepackage{url}            
\usepackage{booktabs}       
\usepackage{amsfonts}       
\usepackage{nicefrac}       
\usepackage{microtype}      
\usepackage{lipsum}
\usepackage{fancyhdr}       
\usepackage{graphicx}       

\RequirePackage{algorithm}
\RequirePackage{algorithmic}
\RequirePackage{eso-pic} 
\RequirePackage{forloop}

\usepackage{microtype}
\usepackage{graphicx}
\usepackage{subfigure}
\usepackage{booktabs} 

\usepackage{listings}
\usepackage{xcolor}

\usepackage{amsmath}
\usepackage{amssymb}
\usepackage{mathtools}
\usepackage{amsthm}

\usepackage{pythonhighlight}

\usepackage{textcomp}
\usepackage[capitalize,noabbrev]{cleveref}

\usepackage[textsize=tiny]{todonotes}
\theoremstyle{plain}
\newtheorem{theorem}{Theorem}[section]

\theoremstyle{definition}

\theoremstyle{remark}

\title{Online Pseudo-average Shifting Attention(PASA) for Robust Low-precision LLM Inference: Algorithms and Numerical Analysis
\thanks{\textit{\underline{Corresponding Author}}}
}

\author{
  Long Cheng*, Qichen Liao \\
  Huawei Technologies Co., Ltd. \\
  China \\
  \texttt{chenglong86@huawei.com} \\
   \And
  Fan Wu \\
  Xiamen University\\
  China \\
  \texttt{wfanstory@stu.xmu.edu.cn}
   \And
  Junlin Mu \\
  Beijing Jiaotong University\\
  China \\
  \texttt{mujunlin@bjtu.edu.cn} \\
    \And
  Tengfei Han \\
  Tohoku University \\
  Japan \\
  \texttt{hantengfei013@gmail.com} \\
   \And
  Zhe Qiu \\
  Fudan University \\
  China \\
  \texttt{qiuzhe27@gmail.com} \\
   \And
  Lianqiang Li \\
  Shanghai Jiaotong University \\
  China \\
  \texttt{sjtu\_llq@alumni.sjtu.edu.cn} \\
   \And
  Tianyi Liu \\
  Cambridge Research Institute \\
  Huawei Technologies Co., Ltd. \\
  Cambridge, UK \\
  \texttt{tianyi.liu3@h-partners.com} \\
   \And
  Fangzheng Miao, Keming Gao, Liang Wang, Zhen Zhang, Qiande Yin \\
  Huawei Technologies Co., Ltd. \\
  China \\
  \texttt{miaofangzheng@huawei.com} \\
}

\begin{document}
\maketitle

\begin{abstract}
Attention calculation is extremely time-consuming for long-sequence inference tasks, such as text or image/video generation, in large models. To accelerate this process, we developed a low-precision, mathematically-equivalent algorithm called PASA, based on Flash Attention. PASA introduces two novel techniques: online pseudo-average shifting and global recovering. These techniques enable the use of half-precision computation throughout the Flash Attention process without incurring overflow instability or unacceptable numerical accuracy loss. This algorithm enhances performance on memory-restricted AI hardware architectures, such as the Ascend Neural-network Processing Unit(NPU), by reducing data movement and increasing computational FLOPs. The algorithm is validated using both designed random benchmarks and real large models. We find that the large bias and amplitude of attention input data are critical factors contributing to numerical overflow ($>65504$ for half precision) in two different categories of large models (Qwen2-7B language models and Stable-Video-Diffusion multi-modal models). Specifically, overflow arises due to the large bias in the sequence dimension and the resonance mechanism between the query and key in the head dimension of the Stable-Video-Diffusion models. The \textit{resonance} mechanism is defined as phase coincidence or 180-degree phase shift between query and key matrices. It will remarkably amplify the element values of attention score matrix. This issue also applies to the Qwen models. Additionally, numerical accuracy is assessed through root mean square error(RMSE) and by comparing the final generated texts and videos to those produced using high-precision attention.
\end{abstract}


\section{Introduction}
\label{Introduction}

The performance improvement of inference capabilities in large models is heavily based on the scaling law \autocite{kaplan2020scaling} of effective
computing power and computational efficiency of neural networks. 

Currently, most large models(LM) are designed based on the transformer architectures\autocite{vaswani2017attention}, where the attention mechanism serves
as a core component. However, the computational complexity of the naive attention computation is as high as quadratic in
relation to input sequence length. The fact is that the sequence length is becoming long with the popularity of chain-of-thought(CoT)\autocite{Jaech2024openAI} and multi-modal 
large models\autocite{liu2024sora,lin2024open,jacobs2023deepspeed} in large models(LM). Besides, LM usually works on modern
neural-network processing units(NPU)\autocite{Liao2019AscendNPU} or general-purpose graphic processing units(GPU)\autocite{Choquette2020GPUA100, Choquette2022GPUH100,Choquette2024GPUB200} with hierarchical memories. Complex
data transfer of intermediate variables in attention calculation on these hardwares significantly impacts LM inference
efficiency and latency\autocite{dao2022flashattention,dao2023flashattention2,shah2024flashattention}. It becomes more and more significant to reduce the latency and improve the performance of the attention calculation.

Flash attention(FA)\autocite{dao2022flashattention,dao2023flashattention2,shah2024flashattention} is one of the most effective approach to enhance the computational performance by optimizing kernels without simplifying the calculations. The tiling strategy in FA is utilized to fully parallelize attention computation. 
Unlike sparse attentions\autocite{dao2022monarch} or linear attentions\autocite{gu2023mamba}, the computational complexity of FA is as high as $O(N^2)$ where $N$ is the input sequence length of transformers, but it is still widely accepted due to the higher reliability than linear or sparse attentions in various scenarios like language, image or video generation\autocite{han2024bridging,dao2022monarch}. Contrast to the remarkable advantages of sparse or linear attention in linear complexity of $O(N)$, 
when input sequence lengths become large, the acceleration of FA is highly dependent on two techniques: low-precision computing including 
quantization algorithms and high-efficiency pipelining algorithms on hardwares. In the recent work of FlashAttention-3\autocite{shah2024flashattention}, the FP8 version of FA is put forward on H100 GPUs.
Due to the well support of FP8 and FP16 format on Tensor Cores(TC), approximately 740 TFLOPS and 1.2PFLOPS are reached for FP16 and FP8, separately. The utilization ratio for the peak performance is about $75\%$.
Besides, it is worth noting that the numerical error in the FlashAttention-3\autocite{shah2024flashattention} with FP8 quantization is quite low thanks to the scaling operation on base 2. However, due to the limited range in FP8 format, the underflow phenomenon usually appears to leading to accuracy loss. To reduce the underflow in FlashAttention-3\autocite{shah2024flashattention}, two techniques are provided to minimize the maximum value in the attention matrix calculation. The first one is 
the block quantization which divides the query matrix into multiple blocks to get the maximum value. The other is called incoherent processing, which introduces two orthogonal random matrices to reduce the influence of outliers before 
quantizing to low precision. The two techniques help reduce the root mean square error(RMSE) from $3.2e^{-4}$ and $2.4e^{-2}$ to $1.9e^{-4}$ and $9.1e^{-3}$ in FP16 and FP8, separately, in the simulation experiments. 

Compared to the reduction of the underflow influence, recent work\autocite{zhang2024sageattention} shows that the non-zero bias in the key of the input tensor in FA causes a non-negligible effect on the accuracy of the results. The influence 
principle of the large bias can be explained using backward error analysis firstly proposed by Wilkinson\autocite{wilkinson1965rounding,higham2002accuracy}. In the work\autocite{higham2020sharper}, the theoretical analysis and experiments on probabilistic backward error illustrate a fact that the relative accumulation error tends to deteriorate with the bias value increasing in summation, inner-product and matmul operations. The analysis suggests that accumulation error tends to decrease if transforming the vectors or matrices to have zero bias before corresponding operations. In SageAttention\autocite{zhang2024sageattention}, the method is utilized by subtracting the bias value from the key matrices of attention input. The experiments in SageAttention\autocite{zhang2024sageattention} illustrate that the accuracy is remarkably improved in some neural networks. It is also claimed that the overhead from the deduction of the bias value in key matrix is as negligible as only $0.2\%$ on GPU. By contrast, NPU are typically domain-specific architectures(DSA) for AI acceleration. It usually behaves excellent for matrix computation, but with a normal capability of vectorization computation. In addition, the bias value computation along the whole sequence in low precision usually means a huge rounding error\autocite{blanchard2020class}. The global operation for the bias subtraction also fails to utilize the high pipelining efficiency and operator merging opportunities for the online algorithm in FA. Hence, the low overhead result is most likely not suitable for long-sequence scenarios and other AI chip architectures like Nerual-network processing units(NPU). 

In our work, we proposed an online algorithm called PASA(pseudo-average shifting attention) for low-precision attention calculation with a robust numerical stability for different LMs. Different with the previous work\autocite{zhang2024sageattention, shah2024flashattention}, the translation invariance for the softmax calculation\autocite{blanchard2021accurately} is utilized in an online block manner. It not only improves the locality of 
the operation naturally supporting multiple-pipeline optimization, but also has a potential benefit to the reduction of numerical error by shifting the local block bias close to zero. Besides, the multi-step implementation including calculating, subtracting the average value and scaling the data distribution is totally transformed to an efficient batched matmul operation. The advantage makes it possible to fully utilize the matrix engines like CUBE in NPU or tensor cores(TC) in GPU. It is also illustrated that PASA is compatible to the basic procedures of FA, which is important for the reuse of previous optimization empiricism on specific architectures. Finally, the algorithm is validated on randomly generated benchmark cases and real large model tasks including language and video generation. Numerical results are given to show the numerical accuracy and performance improvement compared to high-precision FA operators.

\subsection{Flash Attention}
In attention calculation, the basic inputs include query, key and value matrices from embedding results. These three matrices are represented by $\mathbf{Q} \in \mathbf{R}^{S_1 \times d}$, 
$\mathbf{K}\in \mathbf{R}^{S_2 \times d}$ and $\mathbf{V} \in \mathbf{R}^{S_2 \times d}$ for single batch and head attention situations. The size $S_1$ and $S_2$ are input sequence lengths. For self-attention whose query, key and value matrices are from same input tokens, so, $S_1 = S_2$. By contrast, they are usually unequal for cross-attention calculation widely adopted in multi-modal LMs. Besides, $d$ denotes the head dimension of the input matrices. 

The standard attention calculation in the prefilling phase follows the four steps: (1) first matmul - $\mathbf{S} = \mathbf{Q}\mathbf{K}^T \in \mathbf{R}^{S_1 \times S_2}$; (2) static scaling - $\mathbf{S} = \mathbf{S}/\alpha \in \mathbf{R}^{S_1 \times S_2}$; (3) softmax - $\mathbf{P} = softmax(\mathbf{S}) \in \mathbf{R}^{S_1 \times S_2}$; (4) second matmul - $\mathbf{O} = \mathbf{PV} \in \mathbf{R}^{S_1 \times d}$. The $\alpha = 1/\sqrt{d}$ is the static scaling factor. The naive implementation of the attention is time-consuming and not scalable due to the intermediate large matrix - $\mathbf{S}$. It can be solved by introducing blocked algorithms to the above four steps, which is well utilized in FA 1.0\autocite{dao2022flashattention}, 2.0\autocite{dao2023flashattention2} and 3.0\autocite{shah2024flashattention}. The implementation of FA is able to be decomposed into the following fundamental procedures:
\begin{equation}
\mathbf{S}^j_{i} = \mathbf{Q}_i\mathbf{K}_j^T \in \mathbf{R}^{s_1 \times s_2} \label{eq: S=QK}
\end{equation}
\begin{equation}
\mathbf{S}^j_{i} = \mathbf{S}_{i,j}/\alpha  \in \mathbf{R}^{s_1 \times s_2} \label{eq: S=S/afa}
\end{equation}
\begin{equation}
\mathbf{P}^j_{i}, l_j, m_j = softmax(\mathbf{S}^j_{i}) \label{eq: P=softmax(S)}
\end{equation}
where
\begin{equation}
m_j = max(m_{j-1}, rowmax(\mathbf{S}^j_{i})) \in \mathbf{R}^{s_1 \times 1} \label{eq: m=max}
\end{equation}
\begin{equation}
\mathbf{P}^j_{i} = exp(\mathbf{S}^j_{i} - m_j) \in \mathbf{R}^{s_1 \times s_2} \label{eq: P=exp(S-m)}
\end{equation}
\begin{equation}
l_j = exp(m_j - m_{j-1})l_{j-1} + rowsum(\mathbf{P}^j_{i}) \in \mathbf{R}^{s_1 \times 1} \label{eq: l=exp()l+rowsum()}
\end{equation}
Correspondingly, the temporary output can be obtained as:
\begin{equation}
\mathbf{O}_{i} = exp(m_j - m_{j-1})\mathbf{O}_{i} + \mathbf{P}^j_{i}\mathbf{V}_{j}  \in \mathbf{R}^{s_1 \times d} \label{eq: O=exp()O+PV}
\end{equation}
Finally, after sweeping the whole row of the attention matrix, the consequent attention output can be calculated by
\begin{equation}
\mathbf{O}_{i} = \mathbf{O}_{i} / l_{N_{KV}} \in \mathbf{R}^{s_1 \times d} \label{eq: O=O/l}
\end{equation}
Where $s_1 \times d$ and $s_2 \times d$ are the shapes of the basic block for $\mathbf{Q}$ and $\mathbf{KV}$, respectively. Consequently, the total numbers of basic blocks are $N_Q = S_1 / s_1$ and $N_{KV} = S_2 / s_2$. The above variables like $\mathbf{Q}_{i}, (\mathbf{KV})_{i}$ denote the basic block for the original input global matrix. The fundamental idea of FA is to fully utilize the block algorithm to make the calculation in an online method. It enhances the data locality to save memory and improves the pipelining efficiency for the computation-movement overlapping. 
\subsection{Precision Allocations in FA}
\begin{figure}
    \centering
    \includegraphics[width=1\linewidth]{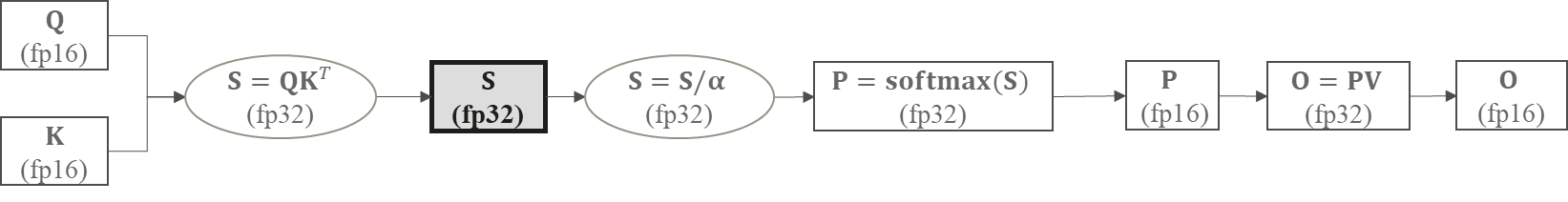}
    \caption{The Precision Allocation in the Original FA}
    \label{fig:Original-FA-Prec}
\end{figure}

Thanks to the powerful matrix engines in AI Chips like NPU CUBE and GPU Tensor Cores(TC), FA is remarkably accelerated by several times to the standard attention. However, the computing power is generally dominated by the low-precision part. For instance, the theoretical performance for FP32 and FP16 Tensor core on Nvidia A100 GPU\autocite{Choquette2022GPUH100, choquette2021nvidia} are almost 156TFLOPS and 312TFLOPS, respectively. By constrast, the numbers are approximately 200TFLOPS and 400TFLOPS for Ascend 910B NPU\autocite{Liao2019AscendNPU,huawei2022Ascend910B}. The original FA 1.0\autocite{dao2022flashattention} and 2.0\autocite{dao2023flashattention2} take the safe allocation method for precision as shown in \cref{fig:Original-FA-Prec}. Except for the input variables on Tensor Cores utilizing FP16 or BF16 precision, the computational procedures including the matmul, scaling, softmax and online update operation are fully FP32 precision. It is worth noting that the numerical overflow or instability almost cannot happen in this precision allocation method due to the large range and high precision of 32-bits floating point numbers. 

\begin{figure}
    \centering
    \includegraphics[width=1\linewidth]{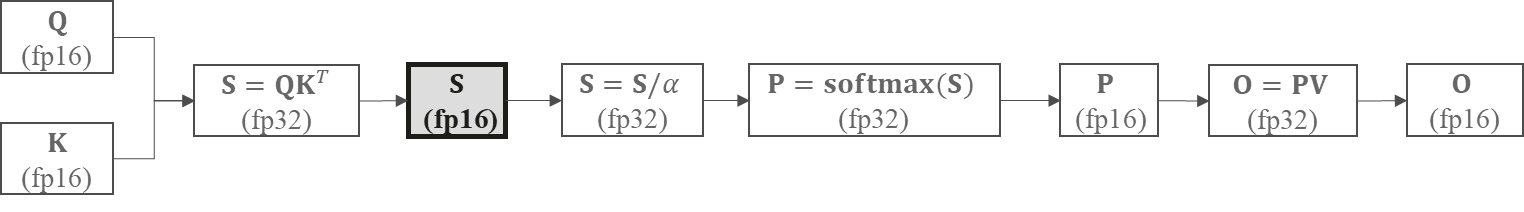}
    \caption{Partially Low Precision(FP16) Allocations in FA}
    \label{fig:Partial_FP16_FA}
\end{figure}

\begin{figure}
    \centering
    \includegraphics[width=1\linewidth]{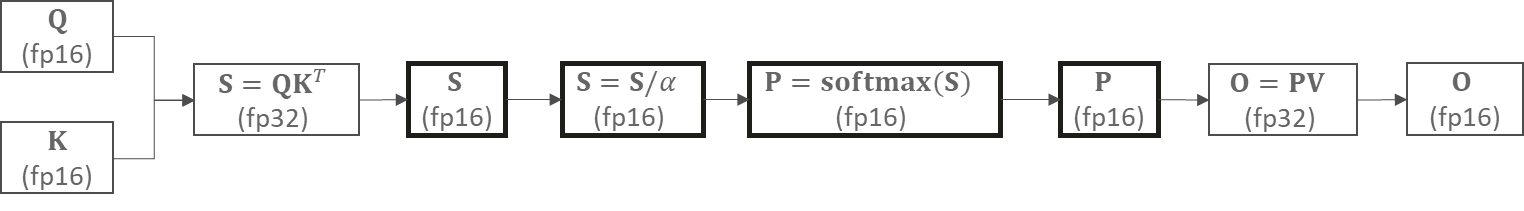}
    \caption{Fully Low Precision(FP16) Allocations in FA}
    \label{fig:Full_FP16_FA}
\end{figure}

In FP3.0\autocite{shah2024flashattention}, the quantization and FP8 are well integrated into the FA 2.0\autocite{dao2023flashattention2} framework for further performance and memory benefits. For NPU-similar architectures, the data movement from the memory L0 near matrix engines to the high-bandwidth memory(HBM) is extremely time-consuming. The adoption of FP32 precision leads to the memory-bound feature of FA in NPU. It has a significant influence on the pipelining efficiency. Hence, it is natural to adjust the precision allocation to the partially low-precision method demonstrated in \cref{fig:Partial_FP16_FA} where the attention score matrix - $\mathbf{S}$ out from the matrix engine is FP16. The two strategies have been adopted in the inference FA operators named \textit{fused\_infer\_attention\_score} of Ascend NPU transformer library\autocite{huawei2022Ascend910B}. Remarkably, the performance discrepancy between the two precision allocation in \textit{fused\_infer\_attention\_score} can be as high as about $(10\% \sim 70\%)$ on Ascend NPU. Hence, the low-precision strategy remarkably releases the memory movement pressure. More aggressively, the full low-precision allocation strategy is illustrated in \cref{fig:Full_FP16_FA} where almost all variables and operations are low precision - FP16. It is expected to be beneficial both for performance and energy consumption.

Nevertheless, the drawback is obvious that low precision means narrow expression range of floating-point numbers as shown in \cref{Tab: Range and Precision}. The consequence is that the overflow phenomenon is more likely to be triggered to introduce \textit{INF}(Infinity Number) in the computation of FA, especially for the diversified input prompts and multiple modals in LM inference phase. Hence, we believe that it is not numerically robust for directly utilizing the low-precision FA framework like the above \cref{fig:Partial_FP16_FA} and \cref{fig:Full_FP16_FA}. In the following part, we will present a new algorithm - PASA, which forms a robust FA framework to support low-precision computing for LM inference especially for mutli-modal LMs like diffusion models. It can maintain a good numerical stability and numerical accuracy for FA computations. 

\begin{table}[t]
\caption{Range and Precision for Different Data Formats.}
\label{Tab: Range and Precision}
\vskip 0.15in
\begin{center}
\begin{small}
\begin{sc}
\begin{tabular}{lcccr}
\toprule
Data Formats & Precision & Overflow Boundary \\
\midrule
FP8       & $6.25\times10^{-2}$   & $448$               \\
FP16      & $4.88\times 10^{-4}$  & $65504$             \\
BP16      & $3.906\times 10^{-3}$  & $3.4 \times 10^{38}$ \\
FP32      & $5.96 \times 10^{-8}$ & $3.4 \times 10^{38}$  \\
\bottomrule
\end{tabular}
\end{sc}
\end{small}
\end{center}
\vskip -0.1in
\end{table}

\section{PASA}
The framework of PASA is compatible with the original FA, so they are mathematically equivalent with each other if not considering numerical rounding error. The difference is that the origin of numerical overflow is eliminated in PASA by adding some mathematical operations in the original procedures of FA. To introduce PASA algorithm, it is essential to present the possible overflow positions in original FA. 
\subsection{The Possibility Analysis of the Appearance of Large Numbers in FA}
It is reasonable to assume that the input tensors - $\mathbf{Q}, \mathbf{KV}$ are within the normal range of low precision data formats such as $65504$ in FP16. The overflow only possibly appears in the computing procedures of Equations \ref{eq: S=QK} to \ref{eq: O=O/l}.

As for the first step in Equation \ref{eq: S=QK}, it is a general-matrix-multiply(GEMM) operation related to inner product, so it possibly holds for the relationship: $\|\mathbf{S}\|_{max} >> \|\mathbf{Q}\|_{max}, \|\mathbf{K}\|_{max}$, which indicates that GEMM could be like an amplifier for original input matrices. By contrast, the second step in Equation \ref{eq: S=S/afa} is like an attenuator due to the static scaling factor - $\alpha$ larger than 1.0. The next phase in Equation \ref{eq: P=softmax(S)} is a softmax operation including Equations \ref{eq: m=max} to \ref{eq: l=exp()l+rowsum()}. The maximum operation cannot generate new large values. The exponent operation in Equation \ref{eq: P=exp(S-m)} is an attenuator because the relationship of $\mathbf{S}^j_{i} - m_i \le 0$ holds. The practical situation is that most of the value in $\mathbf{P} << 1$. Similarly, overflow cannot occur for the Equation \ref{eq: l=exp()l+rowsum()} and \ref{eq: O=exp()O+PV} due to the $rowsum$ and $\mathbf{PV}$ operations conducted in a block manner where the block is usually as small as $128$ or $256$. 

Shortly speaking, the first GEMM operation is the most possible origin if overflow phenomenon appears in the FA computation. 

\subsection{The Mathematical Framework of PASA}\label{sec:Mathematical_Framework}
It is well-known that the softmax operation has a mathematical property of translation invariance\autocite{blanchard2021accurately}. The FA well utilizes the property to maintain numerical stability by subtracting the maximum value in Equation \ref{eq: P=softmax(S)}. In fact, the subtraction operation also can be done in an early stage. It holds for the Equation \ref{eq: softmax(Q(K-K_Mean))} as
\begin{equation}
softmax(\mathbf{Q}\mathbf{K}^T) = softmax(\mathbf{Q}(\mathbf{K}^T - \mathbf{K}_0^T)) \label{eq: softmax(Q(K-K_Mean))}
\end{equation}
where the matrix $\mathbf{K}_0$ is called bias value of the original matrix $\mathbf{K}$. Besides, the component, $\mathbf{k}$, of $\mathbf{K}_0=[\mathbf{k}; ...; \mathbf{k}]$ is same for all rows. In the work\autocite{zhang2024sageattention}, SageAttention is proposed to utilize the above property in Equation \ref{eq: softmax(Q(K-K_Mean))}, and the average vector of the $\mathbf{K}$ matrix along the sequence length direction is utilized as $\mathbf{K}_0$ in Equation \ref{eq: softmax(Q(K-K_Mean))}. The findings are that a large bias in matrix $\mathbf{K}$ is shared by the input tokens, which causes a large numerical error for attention calculation. 
The mathematical framework of PASA also makes use of the property of the invariance in Equation \ref{eq: softmax(Q(K-K_Mean))}. Different with the previous work, the bias is defined as the pseudo-average value of key matrix in an online manner. The corresponding advantage is the locality improvement of the computation. The large bias in the sequence dimension is also the part of the cause of overflow, which will be analyzed in the numerical experiments. In addition, PASA is compatible to current FA frameworks, so it is able to be integrated into online and block quantization algorithms\autocite{shah2024flashattention} as well as recently developed distributed version - ring attention(RA)\autocite{liu2023ring} for multiple devices. 
\begin{figure}
    \centering
    \includegraphics[width=0.8\linewidth]{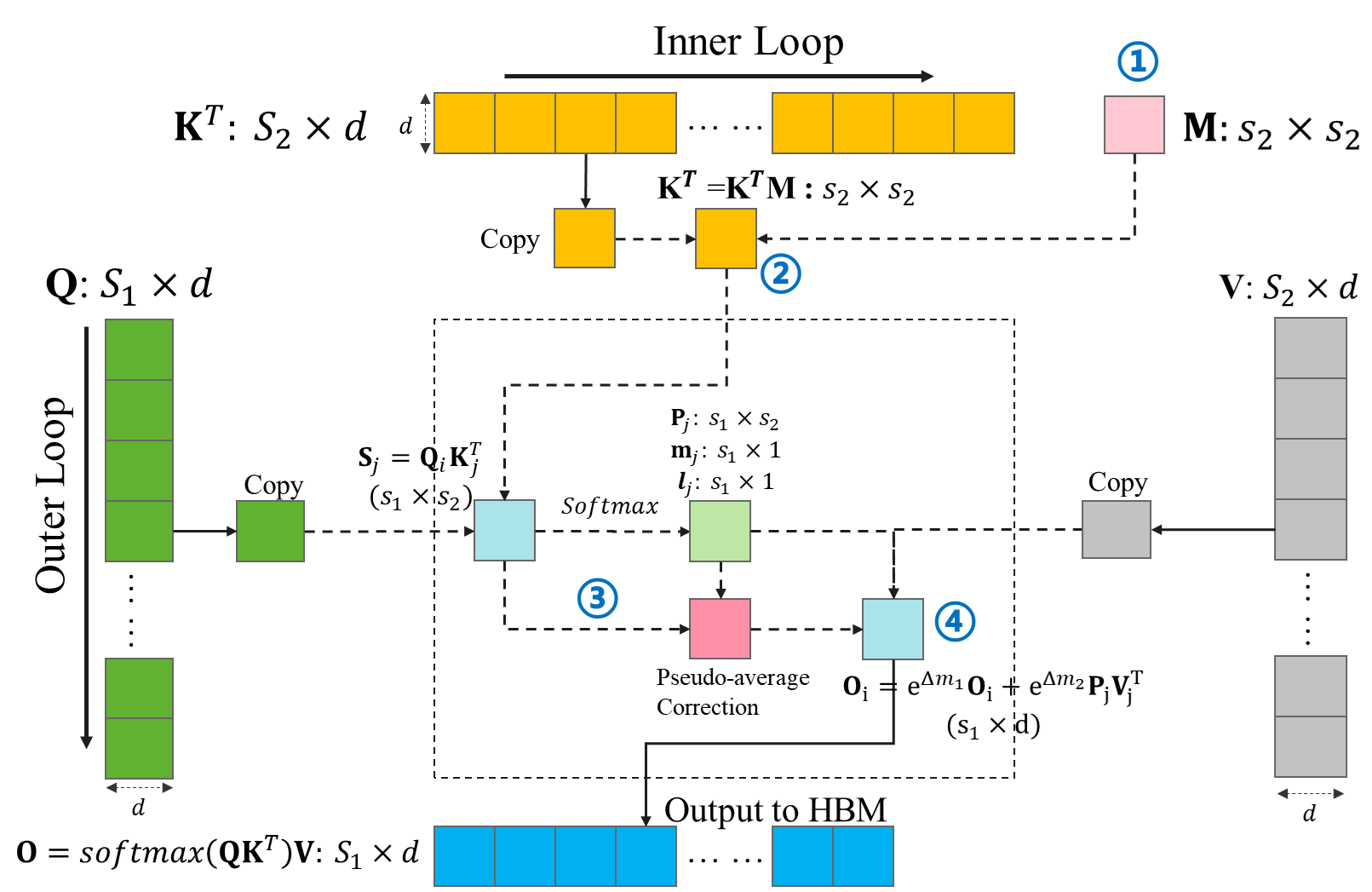}
    \caption{The Diagram Framework of PASA}
    \label{fig:PASA_Framework}
\end{figure}

The schematic diagram for PASA's framework is depicted in Figure \ref{fig:PASA_Framework}. Correspondingly, the whole PASA algorithm is shown in Algorithm \ref{alg:PASA1}. The presented algorithm is naturally suitable to GPU architecture. When it comes to NPU architecture, explicit data movement has to be specified. A special situation is that PASA completely degrades into the FA2.0 algorithm when the hyper-parameter - $\beta$ is set to zero. If the input datatype - \textit{tp} for $\mathbf{Q}, \mathbf{KV}$ is \textit{BF16}, the conversion to \textit{FP16} is needed for PASA algorithm to maintain the optimal accuracy. The fundamental workflow in the Algorithm \ref{alg:PASA1} is similar to the original FA 2.0 except for the added four procedures denoted as "\textcircled{1}\textcircled{2}\textcircled{3}\textcircled{4}". 

\begin{algorithm}[h]
   \caption{PASA Algorithm}
   \label{alg:PASA1}
   \begin{algorithmic}[1]
   \STATE {\bfseries Input:} Matrices $\mathbf{Q}, \mathbf{K}, \mathbf{V}$ (Precision=\textit{tp}) and transformation matrix $\mathbf{M}$ (Precision=\textit{FP16}). Here, $\mathbf{Q}\in \mathbf{R}^{S_1 \times d}, \mathbf{KV} \in \mathbf{R}^{S_2 \times d}, \mathbf{M} \in \mathbf{R}^{s_2 \times s_2}$, and the sub-block matrices of $\mathbf{Q}$ and $\mathbf{KV}$ are $\mathbf{Q}_i \in \mathbf{R}^{s_1 \times d}, \mathbf{KV}_j \in \mathbf{R}^{s_2 \times d}$, while sub-block number are $N_q=S_1/s_1$ and $N_{KV}=S_2/s_2$ with the hyper-parameter - $\beta$ determined by optimal accuracy condition.
   \STATE {\bfseries Output:} the output $\mathbf{O} \in \mathbf{R}^{S_1 \times d}=[\mathbf{O}_1, …,\mathbf{O}_{N_q}]$ with the sub-block matrix $\mathbf{O}_i \in \mathbf{R}^{s_1 \times d}$.
   \STATE Construct the Shifting Matrix - $\mathbf{M}$ for the Treatment of Matrix - $\mathbf{K}$.
   \STATE Initialize the Maximum vector: $m_0=\mathbf{0} \in \mathbf{R}^{s_1 \times 1}, l_0=\mathbf{0} \in \mathbf{R}^{s_1 \times 1}$. (Precision=\textit{FP16})
   \FOR{$j=1$ {\bfseries to} $N_{KV}$}
   \STATE (Batched-GEMM)Pre-processing: $\mathbf{K}^T_j=\mathbf{K}^T_j \mathbf{M} \in \mathbf{R}^{d \times s_2}$. (Precision=\textit{FP16})
   \ENDFOR
   \FOR{$i=1$ {\bfseries to} $N_q$}
   \STATE Initialize the output $\mathbf{O}_i = \mathbf{0} \in \mathbf{R}^{s_1 \times d}$. (Precision=\textit{FP16})
   \FOR{$j=1$ {\bfseries to} $N_{KV}$}
   \STATE (GEMM)Compute Attention Matrix: $\mathbf{S}^j_i=\mathbf{Q}_i \mathbf{K}^T_j \in \mathbf{R}^{s_1 \times s_2}$. (Precision=\textit{FP16})
   \STATE Compute Softmax: $m_j^\prime=rowmax(\mathbf{S}^j_i) \in \mathbf{R}^{s_1 \times 1}, \mathbf{P}^j_i=exp(\mathbf{S}^j_i-m_j^\prime) \in \mathbf{R}^{s_1 \times s_2}, l_j^\prime=rowsum(\mathbf{P}^j_i) \in \mathbf{R}^{s_1 \times 1}$. (Precision=\textit{FP16})
   \STATE Compute Pseudo-average Value: $\bar{\mathbf{S}}^{\prime j}_i=rowmean(\mathbf{S}^j_i) \in \mathbf{R}^{s_1 \times 1}$. (Precision=\textit{FP16})
   \STATE Compute Global pseudo-average Value: $\bar{\mathbf{F}}^{j} = \frac{(j-1) \bar{\mathbf{F}}^{j-1} + \bar{\mathbf{S}}_i^{\prime j}}{j} \in \mathbf{R}^{s_1 \times 1}$. (Precision=\textit{FP16})
   \STATE Compute the Correction Terms of Maximum: $\Delta m^\prime_{j-1} = \frac{\beta(\bar{\mathbf{F}}^{j-1} - \bar{\mathbf{F}}^{j})}{1-\beta}$, $\Delta m^\prime_{j} = \frac{\beta(\bar{\mathbf{S}}_i^{\prime j} - \bar{\mathbf{F}}^{j})}{1-\beta}$. (Precision=\textit{FP16})
   \STATE Compute the Maximum: $m_j = max(m_{j-1} + \Delta m^\prime_{j-1}, m^\prime_{j} + \Delta m^\prime_{j}) \in \mathbf{R}^{s_1 \times 1}$. (Precision=\textit{FP16})
   \STATE Compute the Correction Terms: $\Delta m_{j-1} = m_{j-1} - m_j + \Delta m^\prime_{j-1}$, $\Delta m_{j} = m^\prime_{j} - m_j + \Delta m^\prime_{j}$. (Precision=\textit{FP16})
   \STATE Compute and Update variables: $l_j = exp(\Delta m_{j-1})l_{j-1} + exp(\Delta m_{j})l^\prime_{j} \in \mathbf{R}^{s_1 \times 1}$. (Precision=\textit{FP16})
   \STATE (GEMM)Compute Temporary Output: $\mathbf{O}_i^{j} = \mathbf{P}^j_i \mathbf{V}_j$. (Precision=\textit{FP16})
   \STATE Update the Output: $\mathbf{O}^j_i = exp(\Delta m_{j})\mathbf{O}^j_i + exp(\Delta m_{j-1}) \mathbf{O}^{j-1}_i \in \mathbf{R}^{s_1 \times d}$. (Precision=\textit{FP16})
   \ENDFOR
   \STATE Update the Final Output: $\mathbf{O}_i=\mathbf{O}_i^{N_{KV}}/l_{N_{KV}}$. (Precision=\textit{FP16})
   \ENDFOR
   \STATE Return the Output $\mathbf{O}$.
\end{algorithmic}
\end{algorithm}

\textbf{(1) Construction of the Shifting Matrix and Matrix Preprocessing for Steps \textcircled{1} and \textcircled{2}}

The naive implementation of the bias subtraction of matrix $\mathbf{K}$ includes the reduction operation and then the subtraction operation. It is only able to be treated on vector cores which usually has a limited computing power. Besides, the accumulation rounding error is quite high for long-sequence LM inference. In PASA, we propose a matrix-naive method to tackle the bias subtraction on matrix engines like NPU CUBE or GPU Tensor Cores. The shifting matrix is defined as:
\begin{equation}
\mathbf{M} = \frac{\mathbf{I}}{\alpha} - \frac{\beta \mathbf{J}}{\alpha s_2} = \frac{1}{\alpha} \left[
\begin{matrix}
1-\beta/s_2 & ... & -\beta/s_2 \\
\vdots & \ddots & \vdots \\
-\beta/s_2 & ... & 1-\beta/s_2
\end{matrix}
\right] \label{eq: revised shifting matrix}
\end{equation}
where $\alpha = \sqrt{d}$ is the static scaling factor in FA. $\mathbf{I} \in \mathbf{R}^{s_2 \times s_2}$ represents the identity matrix, and $\mathbf{J} \in \mathbf{R}^{s_2 \times s_2}$ is the all-ones matrix. $\beta$ in the Equation \ref{eq: revised shifting matrix} is a super-parameter which is used to control the percentage of the bias subtraction from the original mean value of matrix $\mathbf{K}$. It can be determined using the optimal accuracy condition shown in the following part. It is worth noting that the pseudo-average value is automatically subtracted when the above shifting matrix - $\mathbf{M}$ is applied to the right hand side of any matrices.

\begin{figure}
    \centering
    \includegraphics[width=0.7\linewidth]{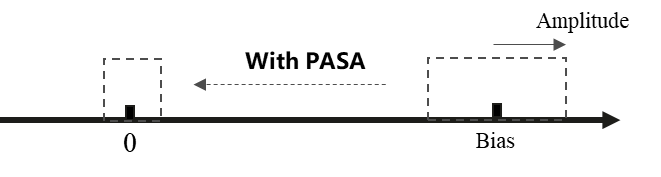}
    \caption{The Diagram for the Reduction of both Average Value and Amplitude with PASA.}
    \label{fig:PASA_Average_Amplitude}
\end{figure}

With the above shifting matrix $\mathbf{M}$, the pseudo-average value calculation, shifting and static scaling of the attention score matrix can be implemented in one step. The consequence is that both the average value and amplitude of the matrix $\mathbf{S}$ are remarkably reduced as depicted in Figure \ref{fig:PASA_Average_Amplitude}. The advantages of merged treatment are helpful to reduce the overhead as well as tackle different transformer architectures in LMs with very large or small average bias. 

Hence, the relationship holds as:
\begin{equation}
\mathbf{K}_j^{{\prime}T} = \mathbf{K}_j^T\mathbf{M} = (\mathbf{K}_j^T - \beta \bar{\mathbf{K}}_j^T)/\beta
\in \mathbf{R}^{d \times s_2} \label{eq: K^T=K^TM}
\end{equation}
\begin{equation}
\mathbf{S}_i^{\prime j} = \mathbf{Q}_i \mathbf{K}_j^{{\prime}T} = (\mathbf{Q}_i \mathbf{K}_j^T)\mathbf{M} = \mathbf{S}_i^{j} \mathbf{M} \in \mathbf{R}^{s_1 \times s_2} \label{eq: S=QK^T M}
\end{equation}
where $\bar{\mathbf{K}}_j = repmat(rowsum(\mathbf{K})/s_2, s_2, 1) \in \mathbf{R}^{s_2 \times d}$ is the average matrix of matrix $\mathbf{K}_j$ along sequence dimension. $rowsum$ is the summation function for the row elements of the matrix. Besides, the Equation \ref{eq: S=QK^T M} is naturally derived from Equation \ref{eq: K^T=K^TM}. It suggests that the subtraction of the matrix $\mathbf{K}_j$ is equivalent to the subtraction of the attention score matrix $\mathbf{S}_i^j$. This is part of explanation that PASA eliminates the generation origin of large values in attention calculation. 

\textbf{(2) Online Recovering of Global Bias Information for Step \textcircled{3}}

The above treatment of subtracting the pseudo-average bias value of the attention score matrix is finished in a local block. It indicates that different values are subtracted in different basic blocks for the attention score matrix. It is subsequently not reasonable to compare the maximum value $m$ and sum up the attention weight matrix $\mathbf{P}$ for different basic blocks.

In the current \textit{i-th} row, the normal softmax operation is conducted as the Equation \ref{eq: P=softmax(S)} on matrix $\mathbf{S}^j_i$ for block $(i,j)$. The uncorrected maximum and summation vectors as well as the attention weight matrix are obtained as $(m_j^\prime, l_j^\prime, \mathbf{P}^j_i)$. From the above analysis, we know that the maximum and summation vectors are not comparable with $(m_j^\prime, l_j^\prime, \mathbf{P}^j_i)$ for the previous block-$(i, j-1)$. Hence, before performing Equation \ref{eq: m=max}, the computation of global correction terms is essential. The theorem for online recovering pseudo-average information is derived as Theorem \ref{thm: Inverse}. 
\begin{theorem}
  Let $\mathbf{I} \in \mathbf{R}^{s \times s}$ represents the identity matrix, and $\mathbf{J} \in \mathbf{R}^{s \times s}$ is the all-ones matrix. $\lambda$ is a parameter not smaller than zero. The shifting matrix is defined as $\mathbf{M} = \mathbf{I} - \lambda \mathbf{J}$. The inverse of the shifting matrix is $\mathbf{M}^{-1} = \mathbf{I} + \frac{\lambda}{1-\lambda s} \mathbf{J}$, where the necessary and sufficient condition of the existence of the inverse is that $\lambda \neq 1$. 
  \label{thm: Inverse}
\end{theorem}

With the help of Theorem \ref{thm: Inverse}, the mean value relationship in Equation \ref{eq: average relationship of matrix S} between the shifted matrix $\mathbf{S}_i^{\prime j}$ and the original matrix $\mathbf{S}_i^{j}$ is able to be derived by multiplying the all-ones matrix $\mathbf{J}$ to the right hand side of the Equation \ref{eq: inverse of matrix S}. The relationship indicates that the original bias value information can be completely recovered if we have the average information of the shifted matrix $\mathbf{S}_i^{\prime j}$.
\begin{equation}
\mathbf{S}_i^{\prime j} (\mathbf{I} + \frac{\beta}{(1-\beta)s_2} \mathbf{J}) = \mathbf{S}_i^{j}
\label{eq: inverse of matrix S}
\end{equation}
\begin{equation}
\frac{\bar{\mathbf{S}}_i^{\prime j}}{1-\beta} = \bar{\mathbf{S}}_i^{j} \in \mathbf{R}^{s_1 \times 1}
\label{eq: average relationship of matrix S}
\end{equation}
where $\bar{\mathbf{S}}_i^{\prime j} = rowsum(\mathbf{S}_i^{\prime j})/s_2$ and $\bar{\mathbf{S}}_i^{j} = rowsum(\mathbf{S}_i^{j})/s_2$ represent the average matrixs for the matrix $\mathbf{S}_i^{\prime j}$ and $\mathbf{S}_i^{j}$, respectively.

The global average matrix for the $j$ blocks needs to be obtained by revising the local matrix $\bar{\mathbf{S}}_i^{\prime j}$ as 
\begin{equation}
\bar{\mathbf{F}}^{j} = \frac{(j-1) \bar{\mathbf{F}}^{j-1} + \bar{\mathbf{S}}_i^{\prime j}}{j} \in \mathbf{R}^{s_1 \times 1}
\label{eq: global maean vector}
\end{equation}
where $\bar{\mathbf{F}}^{j-1}$ represents the global average matrix of the previous $j-1$ blocks. Specifically, for the first block with $j=1$, the relation, $\bar{\mathbf{F}}^{j=1} = \bar{\mathbf{S}}_i^{\prime (j=1)}$, holds. 

By utilizing the global average information in Equation \ref{eq: global maean vector}, we can obtain the correction terms, $\Delta m^\prime_{j-1}$ and $\Delta m^\prime_{j}$, of the maximum value in the Algorithm \ref{alg:PASA1}. Subsequently, the maximum operation is obtained as $m_j$ between the block $(i, j-1)$ and $(i,j)$ in the Algorithm \ref{alg:PASA1}. The correction terms, $\Delta m_{j-1}$ and $\Delta m_{j}$, are also able to be computed. 

\textbf{(3) Online Correction of the Softmax and Output for Step \textcircled{4}}

After obtaining the correction terms, the following softmax summation in the $j-th$ block can be corrected to obtain $l_j$. Similarly, the output for the $i-th$ block is also able to be updated to obtain $\mathbf{O}_i$. When the $i-th$ row is finished, the final output $\mathbf{O}_i$ is updated by using the Equation \ref{eq: O=O/l}.

\subsection{Optimal Accuracy Condition of PASA in Low Precision Computing}

When associated with finite-precision computing like FP16, the mathematical equivalence property of PASA to FA is destroyed. It leads to the fact that we have to consider the numerical rounding error and find the optimal accuracy implementation of PASA. In this work, the optimal accuracy condition for determining $\beta$ is proposed as Equation \ref{eq: optimal accuracy condition}. The principle for the condition on reducing numerical rounding error as well as avoiding overflow is given in the Appendix \ref{APP: Principle of optimal condition to reduce error}. Besides, the derivation of the nonlinear function in Equation \ref{eq: optimal accuracy condition} is given in the Appendix \ref{APP: The Principle of Using Nonlinear Equation}.
\begin{equation}
\frac{\beta}{1-\beta} = f(\beta)
\label{eq: optimal accuracy condition}
\end{equation}
where the nonlinear function $f(\beta)$ is denoted by the Equation \ref{eq: nonlinear function} in Appendix \ref{APP: Principle of optimal condition to reduce error}. The whole nonlinear equation in Equation \ref{eq: nonlinear function} can be solved by utilizing fixed-point iteration method in Equation \ref{eq: Jacobi Iteration} under high precision $cp = FP64$. The initial values for $\beta$ are chosen as ${1-2^{-4}}$, $1-2^{-5}$ and ${1-2^{-6}}$. The final solution for Equation \ref{eq: optimal accuracy condition} are ${\beta=0.937500}$, $0.968994$ and $0.984497$. We will adopt $\beta = 0.984497$ in the following validation.

\section{Experimental Validation}
The validation for PASA algorithm is conducted on the datasets generated both by random data generators and real LMs. The benefits of PASA on avoiding overflow and reducing numerical error in low precision are well analyzed. 
\subsection{Benchmark Cases}
To include the influence of outlier in the recently published FA 3.0\autocite{shah2024flashattention}, both the uniform random distribution and the hybrid random distribution are considered as depicted in Table \ref{Tab: experiments on Real LMs}. The random cases are generated by \textit{torch.rand}, \textit{torch.norm} and \textit{numpy.random.binomial} in Equations \ref{eq: generation formulation for uniform random} and \ref{eq: generation formulation for hybrid random}. The real LM benchmark cases include large language models and multi-modal large models. 
\begin{equation}
\mathbf{Q}, \mathbf{K}, \mathbf{V} = U(x_0 - Am , x_0 + Am),
\label{eq: generation formulation for uniform random}
\end{equation}
\begin{equation}
\mathbf{Q}, \mathbf{K}, \mathbf{V} = N(x_0, 1) + N(0, Am^2) * Bernoulli(p)
\label{eq: generation formulation for hybrid random}
\end{equation}
where $U(a, b)$ represents uniform random distribution between the range of $[a = x_0 - Am, b = x_0 + Am]$ for the mean value and the amplitude as $x_0$ and $Am$, separately. $N(\mu, \sigma^2)$ is the normal random distribution with the mean value and standard deviation as $\mu = x_0$ and $\sigma = Am$, respectively. Besides, the term - $Bernoulli(p)$ represents the Bernoulli distribution with the probability as $p = 0.001$.  
\begin{table}[t]
\caption{Validation Benchmark Cases for PASA: Random Cases and Real Large Models(Qwen\autocite{bai2023qwen,yang2024qwen2} and Stable-Video-Diffusion\autocite{blattmann2023stable}).}
\label{Tab: experiments on Real LMs}
\vskip 0.15in
\begin{center}
\begin{small}
\begin{sc}
\begin{tabular}{lcc}
\toprule
Large Models         & Category      \\
\midrule
Uniform Random     & Random   \\
Hybrid Random      & Random   \\
Qwen2-7B           & Language         \\
stable-video-diffusion-img2vid         & Multi-modal         \\
\bottomrule
\end{tabular}
\end{sc}
\end{small}
\end{center}
\vskip -0.1in
\end{table}

The metic to measure the effect of avoiding overflow is quite straightforward by observing the appearance of the \textit{INF} or \textit{NAN} in the calculation. For numerical error, we used the relative root-mean-square error(RMSE) to measure the accuracy of PASA. This metic is defined in Equation \ref{eq: RMSE definition}.
\begin{equation}
RMSE = \frac{\| O_{computed} - O_{Golden} \|_2}{\| O_{Golden} \|_2}
\label{eq: RMSE definition}
\end{equation}
where the denominator is used to normalize the output value for different input data magnitudes. More practically, we also present the generated video comparison with the reference one to show the influence of PASA on real inferences. 

\subsection{Validation Platform}
The validation is carried out on PyTorch/Torch-NPU\autocite{huawei2022Ascend910B} in the eager mode. It is quite efficient to validate the accuracy of the algorithm prototype on this platform. To illuminate the advantages of PASA, both the high-performance and high-precision versions of PFA in CANN are considered into the reference group. 

\subsection{Numerical Accuracy Validation}
The input shape of matrices - $\mathbf{Q}$ and $\mathbf{KV}$ for random benchmark cases is kept as $(B, N, S, D) = (1, 16, 1280, 128)$. For real LM cases, the input shapes are determined by LM network structures and input matrices. In our current validation LMs including Qwen\autocite{bai2023qwen,yang2024qwen2} and stable-video-diffusion(SVD) models\autocite{blattmann2023stable}, the typical sequence length $S$ is about $5k \sim 8k$.
\subsubsection{Case 1: Benchmark Datasets for Random Distribution}
As is mentioned in Section \ref{sec:Mathematical_Framework} and Figure \ref{fig:PASA_Average_Amplitude}, the benefit of introducing PASA framework is to reduce both the bias value and the amplitude of attention score matrix. In the experiments related to the random data distribution, we give the random data generation with different mean value and amplitudes to illuminate the influence mechanisms to numerical error and overflow. We present the overall results in Appendix \ref{APP: Numerical Results for Random Cases} for two categories of random data distribution(uniform distribution in Equation \ref{eq: generation formulation for uniform random} and hybrid distribution in Equation \ref{eq: generation formulation for hybrid random}). The detailed description for the numerical behaviors is illustrated in Appendix \ref{APP: Numerical Results for Random Cases}. Only the findings are summarized here. For the uniform data distributions, the results indicate: \par
\quad (1) both the bias/mean value and the amplitude can incur the overflow when directly reducing the computational precision to low precision-FP16 in FA;\par
\quad (2) PASA has a better behavior of the overall numerical accuracy than partially low-precision allocation of FA. 

We also obtained the appearance percentages of the \textit{NAN} element in \cref{Tab: Statistics for the NAN Percentages-high-per FA} located in \cref{APP: Statistics for the NAN Percentages}. Contrast to the full \textit{NAN} value in the attention output for a large mean value $x_0 = 30$ in the uniform data distribution, only a small percentage about $0.12\%\sim8.14\%$ of the attention output is \textit{NAN}, which still leads to the inference failure in LMs. 

Furthermore, the findings for hybrid distribution are consistent with the results in the above uniform distribution. The sources of overflow or \textit{NAN} include the large mean value and the large amplitude of the input query, key matrices. The two factors make it difficult to directly reduce the precision to FP16 like the algorithm of partially low-precision allocation of FA(FP16-FP32). To verify whether real LMs also have the above phenomena, we conducted numerical experiments on real LMs as shown in the following part.

\subsubsection{Case 2: Real Large Models}
\begin{figure}
    \centering
    \subfigure[Qwen2-7B(Original)]{
        \includegraphics[width=0.45\linewidth]{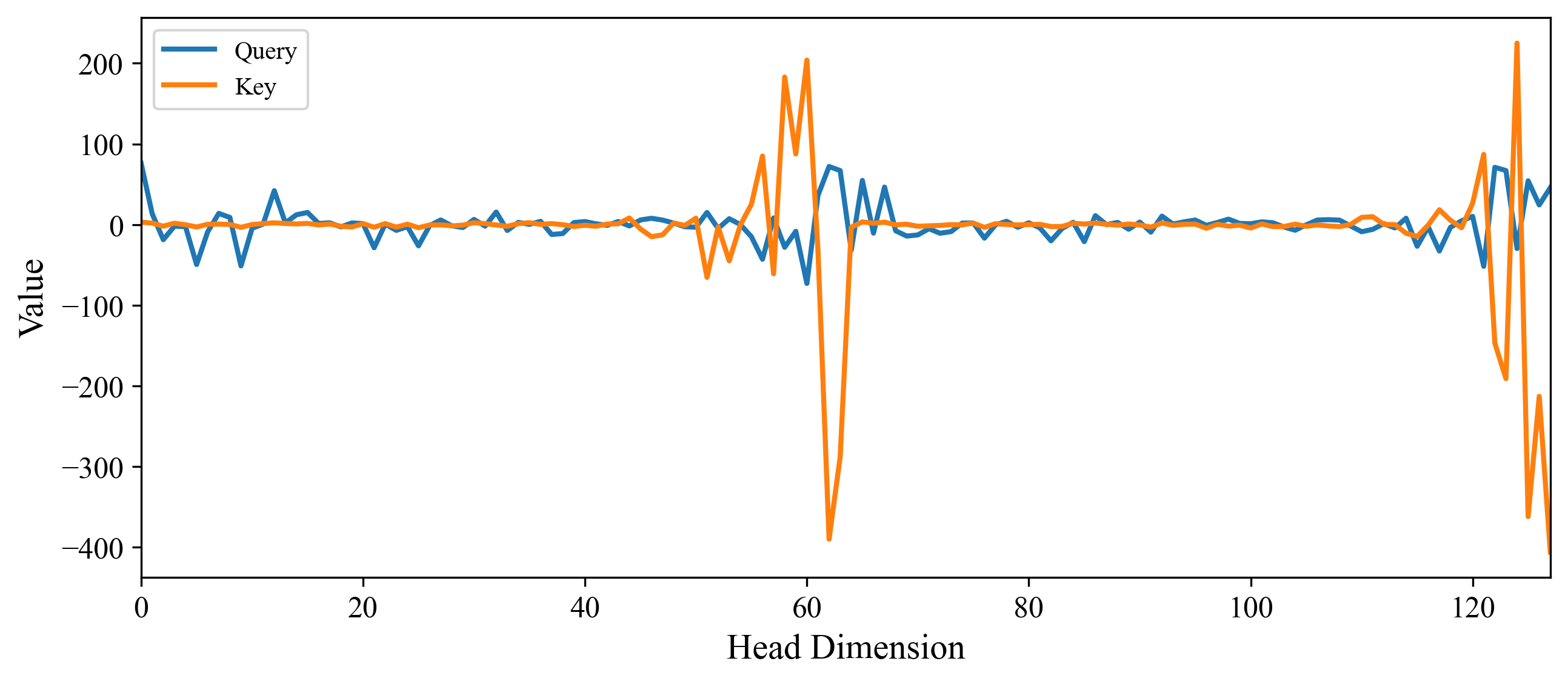}
    }
    \subfigure[IMG2VID(Original)]{
        \includegraphics[width=0.45\linewidth]{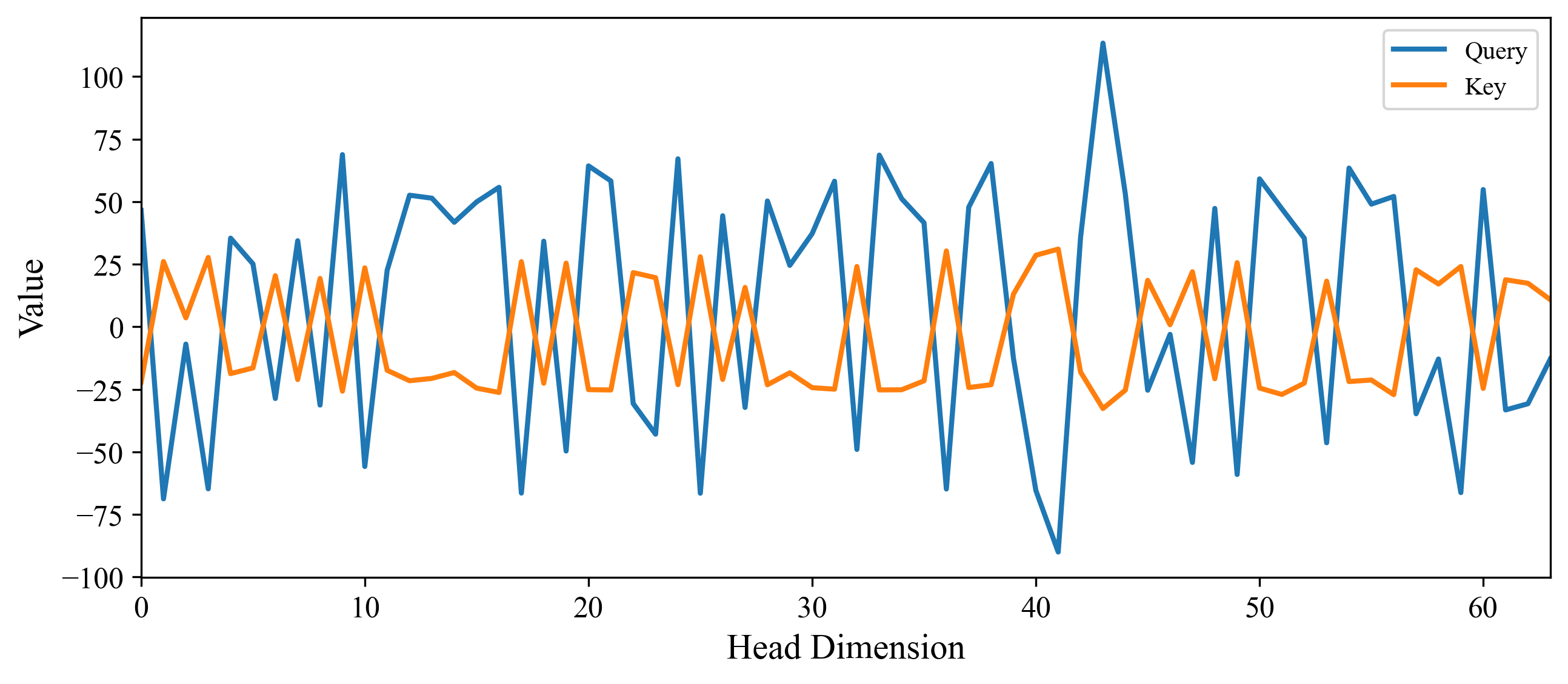}
    }
    \subfigure[Qwen2-7B(With PASA)]{
        \includegraphics[width=0.45\linewidth]{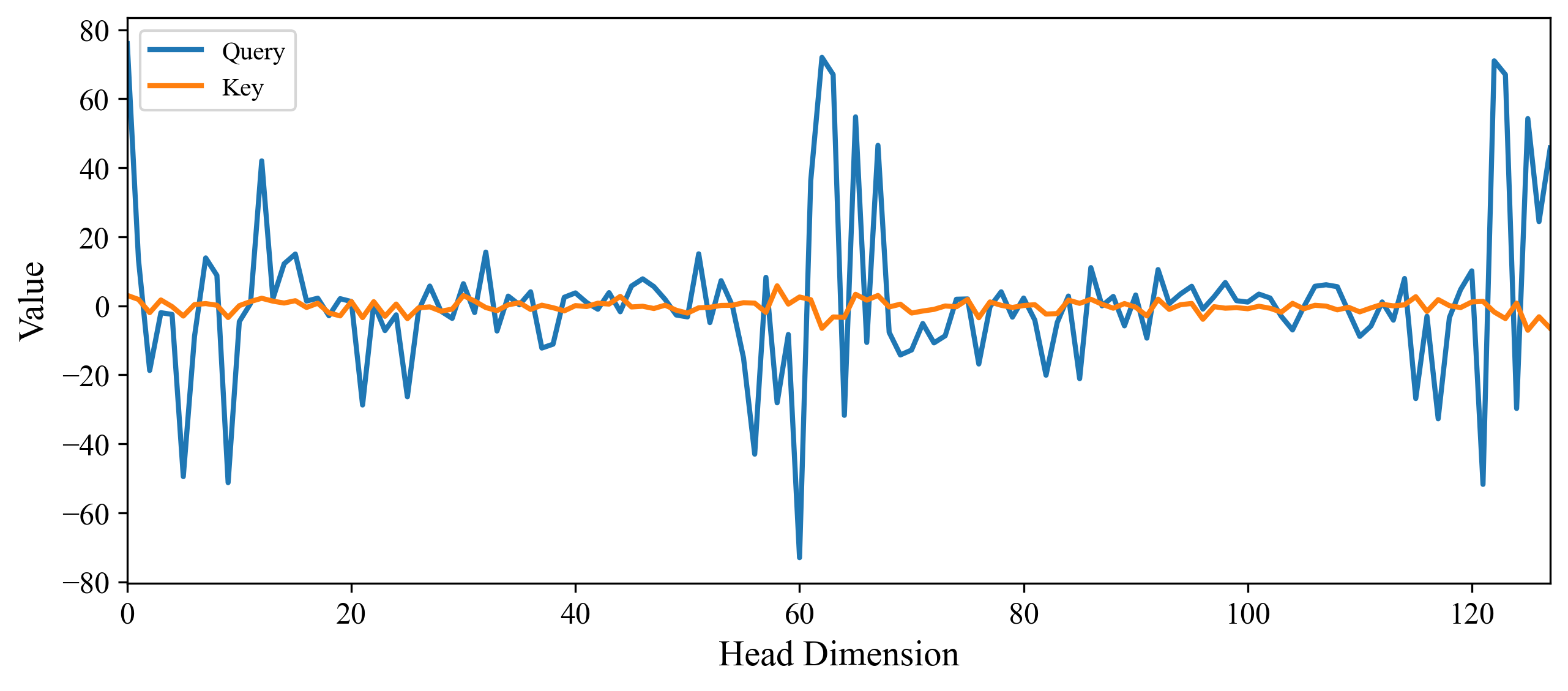}
    }
    \subfigure[IMG2VID(With PASA)]{
        \includegraphics[width=0.45\linewidth]{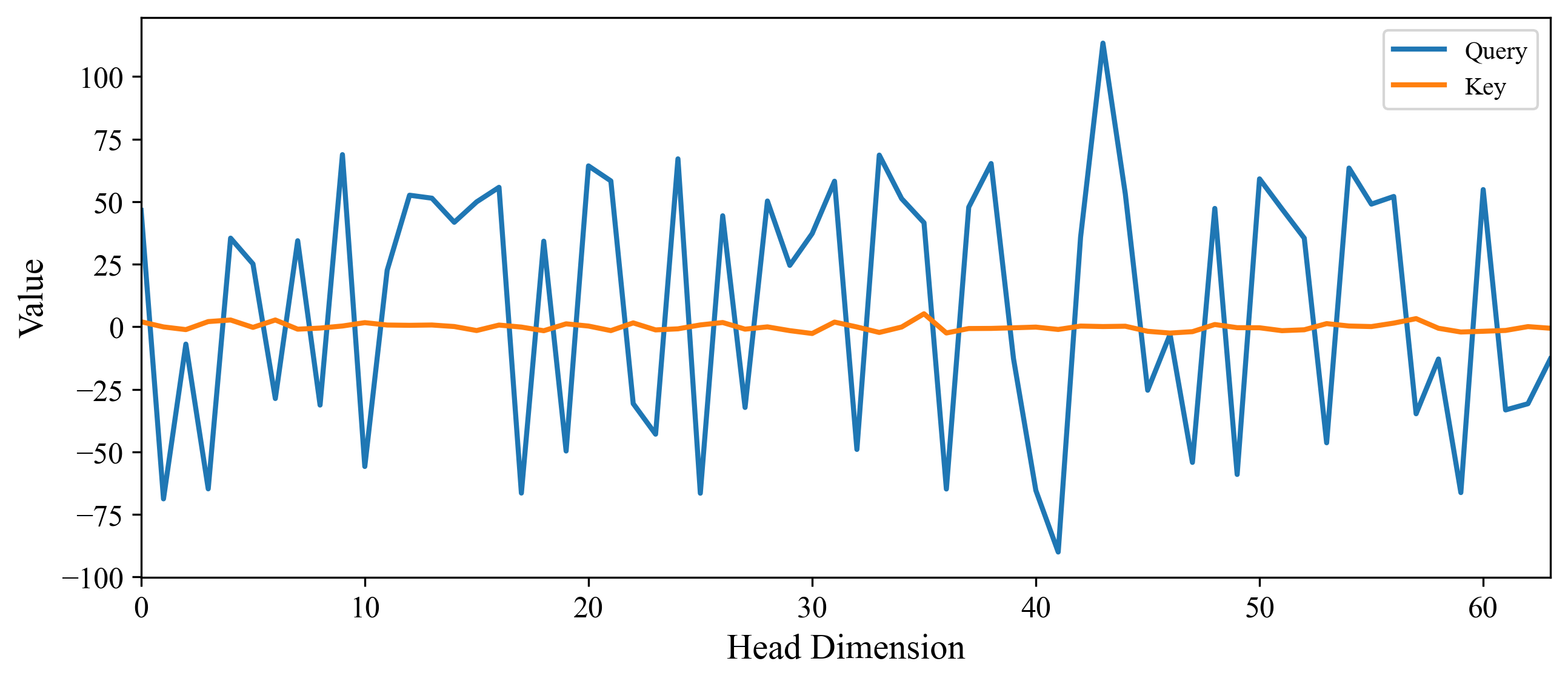}
    }
    \caption{The Sampling Data Distribution for Query and Key in Different Dimensions for Multi-modal LMs(Stability-AI/stable-video-diffusion) for the Overflow Cases.}
    \label{fig:Data Distribution for Query and Key-SVD}
\end{figure}

To verify the overflow mechanism in real LM cases, we conducted the experiments on typical open-source LMs as depicted in Table \ref{Tab: experiments on Real LMs}. We find the overflow cases of attention calculation by instructing sample code. The code checks whether the matmul result of $\mathbf{Q}\mathbf{K}^T$ exceeds the maximum normal value - $65504$ in \textit{FP16} precision. In the current found overflow cases, the shapes for the Query, Key and Value matrices are $[Batch, Head, Seq\_len, Dim] = [1, 28, 5676, 128]$ and $[50, 5, 9216, 64]$ for \textit{Qwen} and \textit{IMG2VID} models, respectively. In the Figures \ref{fig:cloud pictureS for The original QK and preprocessed K in Qwen2-7B}(a-b) and \ref{fig:cloud pictureS for The original QK and preprocessed K in IMG2VID}(a-b) of Appendix \ref{APP: Cloud Maps for Real LMs}, the cloud maps are depicted for the $\mathbf{QK}$ matrices. It is observed that intensive oscillation occurs along the head dimension of the original $\mathbf{QK}$ matrices. Meanwhile, remarkable bias is shown along the sequence dimension of the original $\mathbf{QK}$ matrices. Nevertheless, after the pre-processing with PASA, the data range is massively reduced from $[-412.0, 234.0]$ to $[-12.54, 9.976]$ and from $[-34.44, 33.88]$ to $[-4.283, 5.843]$ for Key matrices in Qwen2 model and IMG2VID model, respectively. It further reduces the data range of attention score matrices from $[-226360, 27757]$ to $[-58134,1124]$ and from $[-86569, -67503]$ to $[-3402, 1752]$ for the two models, respectively. These range of data can be well represented by FP16 data format without suffering from overflow. 
Moreover, the data from the center line in the sequence dimension for the original $\mathbf{QK}$ Matrices are plotted in Figures \ref{fig:Data Distribution for Query and Key-SVD}(a) and (b). By contrast, the preprocessed results with PASA are also depicted in Figures \ref{fig:Data Distribution for Query and Key-SVD}(c) and (d). We have the following significant findings:\par
\quad (1) The \textit{resonance} mechanism between Query and Key matrices along the head dimension is the occurrence cause of large values in attention score matrices both for language model(Qwen2) and multi-modal model(IMG2VID);\par
\quad (2) PASA efficiently removes the origin of large values by eliminating the \textit{resonance} amplitude along head dimension. 

\begin{figure}
    \centering
    \subfigure[Category 1: Phase Lag = $180^o$]{
        \includegraphics[width=0.4\linewidth]{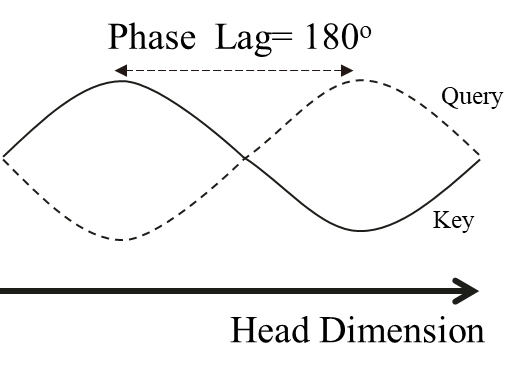}
    }
    \subfigure[Category 2: Phase Lag = $0^o$]{
        \includegraphics[width=0.4\linewidth]{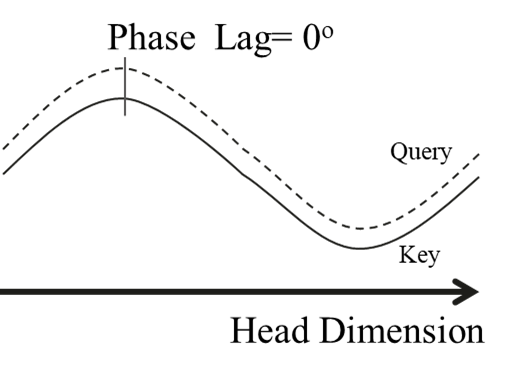}
    }
    \caption{The Definition of \textit{Resonance} in Attention Calculation(The Category 1 will Cause Large Negative Values, while the Category 2 will Cause Large Positive Values).}
    \label{fig:definition of resonance}
\end{figure}

The concept of \textit{resonance} means that amplitude and energy can be extremely amplified in physical systems when the frequency or wave length and phase of external forces are identical to the inherent frequency and phase of a system. In attention calculations of the above LM cases, the query matrices can be viewed as the external forces. The occurrence of \textit{resonance} means that the frequency or the wave length of the query matrices along head dimension are almost identical to these of key matrices. Meanwhile, the phase lag is $180$ degree. It subsequently causes the negative large values which possibly lead to the numerical overflow for half precision due to the inner product operation of $\mathbf{Q}\mathbf{K}^T$. Generally, we give the definition of \textit{resonance} in attention calculation in \cref{fig:definition of resonance} in two categories of situations for attention calculation.

To clearly illuminate the end-to-end effect of PASA on the LM inference, the two generated video clips are illustrated in Figure \ref{fig:SVD Generated Video Clips} for IMG2VID model(The input information is provided in Appendix \ref{APP: Prompt Information for LM Inference Cases}). No any overflow phenomena happen in the whole inference process using IMG2VID with PASA. Particularly, the inference accuracy with PASA is almost same with the reference video quality. The conclusion is also suitable to Qwen2-7B model as shown in Appendix \ref{APP: Prompt Information for LM Inference Cases}. It indicates that it is possible to utilize the half-precision computing in attention for high-quality inference tasks without losing model output accuracy and numerical robustness. 
\begin{figure}
    \centering
    \subfigure[Original Generation]{
        \includegraphics[width=0.44\linewidth]{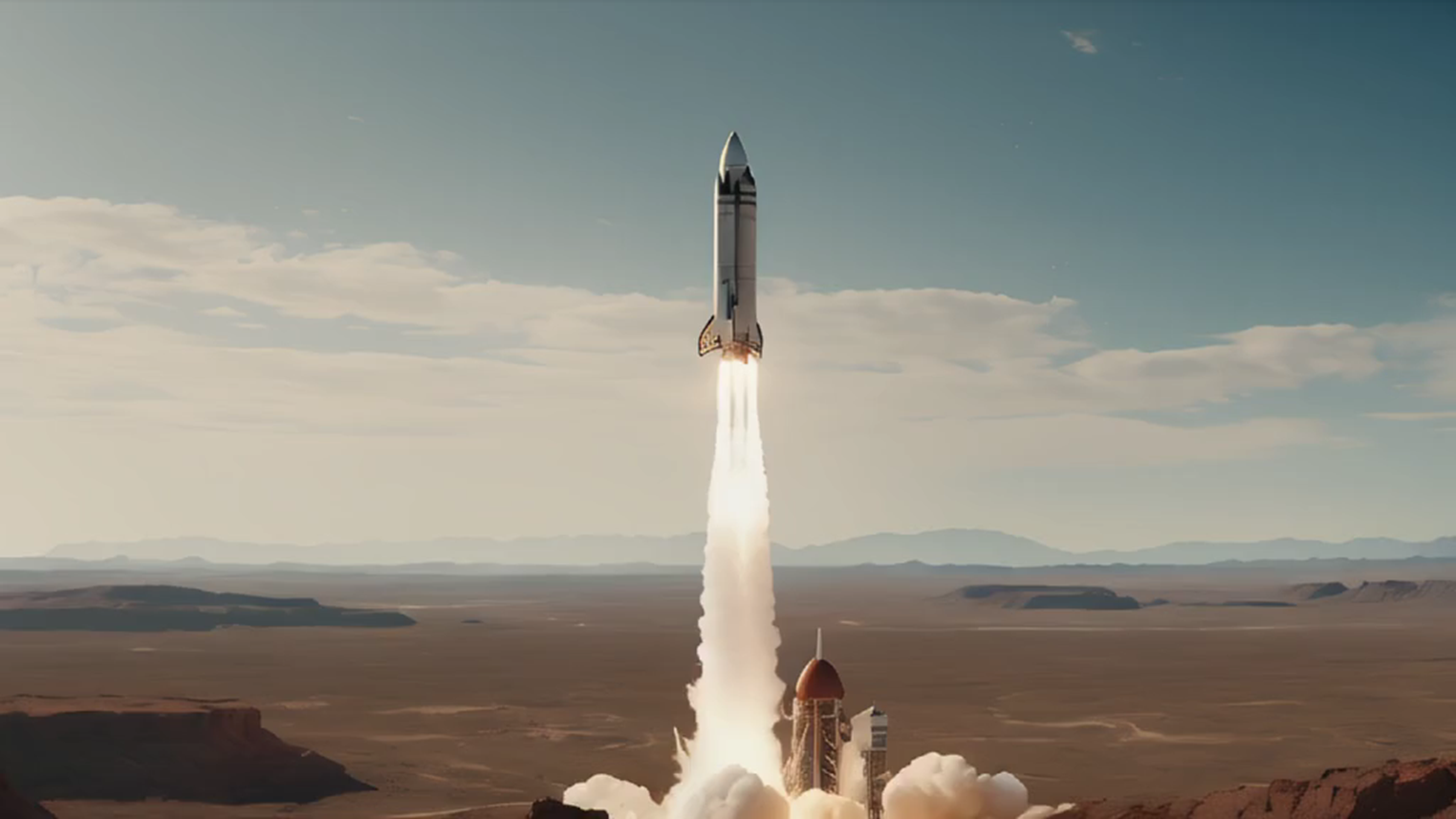}
    }
    \subfigure[Generation with PASA]{
        \includegraphics[width=0.44\linewidth]{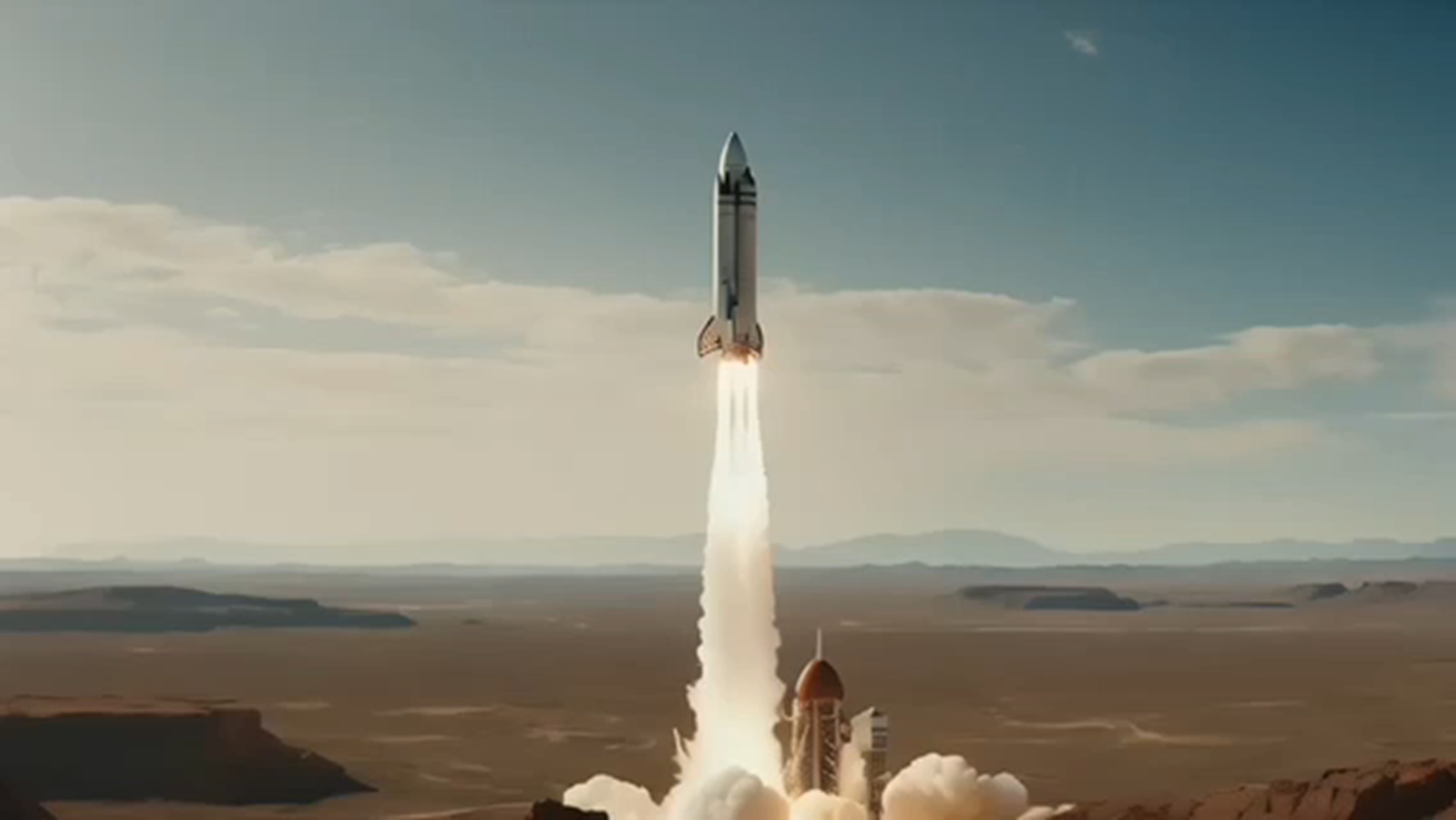}
    }
    \caption{Generated Video Clips with the Original High-precision FA and Low-precision PASA in IMG2VID Model.}
    \label{fig:SVD Generated Video Clips}.
\end{figure}

\section{Conclusion and Future Work}
Long-sequence inference is expected prominent for practical complicated large model servings. Nevertheless, the computational complexity is as high as $O(S^2)$ to the sequence length for transformer architectures. We developed a low-precision algorithm called PASA which is mathematically equivalent to Flash Attention. Different to the previous work, PASA introduces two novel techniques: online pseudo-average shifting and global recovering. These techniques enable the usage of half-precision computation throughout the attention process without overflow occurrence. It, subsequently, releases the heavy burden of memory and data movement on AI chips such as Ascend NPU. PASA is validated using both random benchmarks and real large models. Critical findings are that overflow mainly arises due to the large bias in sequence dimension and the \textit{resonance} mechanism of the query and key matrices in head dimension. Both large language models and multi-modal models have similar overflow features. However, it is still an open question whether the mechanism is a common phenomenon or just coincidence in LMs. It needs more statistical analysis for various LMs in the future work.

The current work is more about the mathematical principle of PASA algorithm and its numerical behaviors. The performance improvement for different architectures like NPU and GPU will come in the next. Besides, since that overflow does not always appear, it is also promising to design an adaptive mechanism to start PASA in the future work. Particularly, SageAttention in \autocite{zhang2024sageattention} suggests a promising memory and performance benefits from FP8 quantization. The combination of the current work with FP8 or int8 quantization in a naively online block manner will be a meaningful direction for further attention acceleration for long sequence inference.

\printbibliography

\newpage
\onecolumn
\appendix
\section{The Principle of using Optimal Accuracy Condition in PASA to Reduce Numerical Error and Overflow}
\label{APP: Principle of optimal condition to reduce error}

To maintain the PASA effect of avoiding overflow, the hyper-parameter $\beta$ is essentially as close as the value 1.0 which represents the full subtraction of the average bias. We know that the inverse of the shifting matrix does not exist if $\beta = 1.0$. The nonexistence of the inverse will lead to the failure to obtain the correction terms. Correspondingly, the values very close to 1.0 can be selected as the $\beta$ candidates like $0.9,0.99,0.999$. 

However, it is observed that the rounding error makes it impossible to fully represent some significant values like $(1-\beta/s_2)$ and 
$(-\beta / s_2)$ in the Equation \ref{eq: revised shifting matrix}. The equivalent effect is that the actually used $\beta$ is changed. Nevertheless, in the correction phase of PASA, the original $\beta$ is still applied to form the correction terms.
\begin{equation}
f(\beta) = \frac{b n}{a(a-bn)} + \frac{1-a}{a}
\label{eq: nonlinear function}
\end{equation}
where the $n$ represents the size of the shifting matrix $\mathbf{M}$ in the Equation \ref{eq: revised shifting matrix}, so $n = s_2$. The parameters, $a$ and $b$, given in the Equation \ref{eq: nonlinear function}, are obtained by rounding off the original parameters in the Equation \ref{eq: revised shifting matrix}. 
\begin{equation}
b = fl_{tp}(\beta / n), a = fl_{tp}(1 - \beta / n) + b,
\label{eq: parameters in the nonlinear function}
\end{equation}
where $fl_{tp}(\cdot)$ represents the rounding operation in low precision $tp$ determined by the data format of input $\mathbf{Q}, \mathbf{K}, \mathbf{V}$. It means that $tp = FP16$ if the data format of the input is $FP16$. In the meanwhile, $tp = BF16$ if the data format of the input is $BF16$. The whole nonlinear equation in the Equation \ref{eq: nonlinear function} can be solved by utilizing fixed-point iteration method in the Equation \ref{eq: Jacobi Iteration} under high precision $cp = FP64$. We define the $Inva = \frac{\beta}{1-\beta}$ and $Inva_1 = \frac{b n}{a(a-bn)} + \frac{1-a}{a}$ as the ideal invariance and the practical invariance under rounding effect, respectively. From the table \ref{Tab: Invariance Difference for initial beta}, the minor numerical error is found in the practical low-precision computing of the Invariance quantity compared to the ideal quantity for the initially given hyper-parameter $\beta$. It is worth noting that the supplemented values $1-2^{-4}=0.9375,2^{-5}=0.96875,2^{-6}=0.984375$ are specially chosen because they can completely denoted by the FP16 format without any numerical loss. The minor error of the invariance will cause the aliasing error in the maximum value comparison in the equation \ref{eq: m=max}, which is the dominant error source to the final output of FA. Nevertheless, if $\beta$ is solved by using Equation \ref{eq: Jacobi Iteration} with the initial values, the results given in the Table \ref{Tab: Invariance Difference for optimized beta} illuminate that the invariance error is reduced to zero. The next part will show the end-to-end influence of the optimized hyper-parameter $\beta$. According to the above analysis, it is observed that the parameter $\beta = 0.9375$ is particularly good. It can be fully represented by FP16, and the invariance is an integer. These confirm that no rounding error is caused by the invariance itself in the correction steps of PASA. 

\begin{equation}
\beta_{k+1} = \frac{f(\beta_k)}{1 + f(\beta_k)}
\label{eq: Jacobi Iteration}
\end{equation}

\begin{table}[H]
\caption{Invariance Parameters under Initial and Optimized $\beta$ for the Computing Precision as $FP16$($Inva = \frac{\beta}{1-\beta}$, $Inva_1 = \frac{b n}{a(a-bn)}$, $Rel. Err = \frac{|Inva - Inva_1|}{|Inva|}$).}
\label{Tab: Invariance Difference for initial beta}
\vskip 0.15in
\begin{center}
\begin{small}
\begin{sc}
\begin{tabular}{ccccr}
\toprule
Initial $\beta$   & $Inva$ & $Inva_1$ & Rel. Err. \\
\midrule
$0.9$           & $9.000$    & $8.971$  &   $0.32\%$ \\
$1-2^{-4}$        & $15.00$    & $15.00$  &   $0.0\%$ \\
$1-2^{-5}$       & $31.00$    & $31.25$  &   $0.81\%$ \\
$1-2^{-6}$      & $63.00$    & $63.50$  &   $0.79\%$ \\
$0.99$          & $99.00$    & $102.2$  &   $3.23\%$ \\
$0.999$         & $999.0$     & $1031$   &   $3.20\%$ \\
\bottomrule
\end{tabular}
\hspace{4\parindent}
\begin{tabular}{ccccr}
\toprule
Optimized $\beta$   & $Inva$ & $Inva_1$  & Rel. Err. \\
\midrule
$0.9$           & $8.971$    & $8.971$  &   $0.0\%$ \\
$0.9375$        & $15.00$    & $15.00$  &   $0.0\%$ \\
$0.96899$       & $31.25$    & $31.25$  &   $0.0\%$ \\
$0.984497$      & $63.50$    & $63.50$  &   $0.0\%$ \\
$0.990311$      & $102.2$    & $102.2$  &   $0.0\%$ \\
$0.999031$      & $1031$     & $1031$   &   $0.0\%$ \\
\bottomrule
\end{tabular}
\end{sc}
\end{small}
\end{center}
\vskip -0.1in
\end{table}

\newpage
\section{The Principle of Using Nonlinear Equation to Obtain the Optimal Accuracy Condition.}
\label{APP: The Principle of Using Nonlinear Equation}

The numerical error could be caused by the rounding operation of shifting matrix  - $\mathbf{M}$. As is mentioned above, the exact parameter, $\beta$, and the inverse of the rounded shifting matrix are simultaneously utilized in the correction procedure. If the rounding error is taken into consideration, the inverse of the shifting matrix is not like the Eq. \ref{thm: Inverse}. Correspondingly, the following form also holds for the rounded shifting matrix.
\begin{equation}
\mathbf{M}_{fp} = a \mathbf{I} - b \mathbf{J}
\label{eq: rounded matrix S}
\end{equation}
where $b = fl(-\mathbf{M}[0, 1])$ and  $a = fl(\mathbf{M}[0, 0]) + b$, and $fl(\cdot)$ denotes the rounding operation using FP16.
The general inverse form of the above \cref{eq: rounded matrix S} can be written as the \cref{eq: general inverse of rounded matrix S}.
\begin{equation}
\mathbf{M}'_{fp} = \frac{1}{a} \mathbf{I} - \frac{b}{a(a-bs)} \mathbf{J}
\label{eq: general inverse of rounded matrix S}
\end{equation}
With the help of the \cref{eq: average relationship of matrix S}, it is natural to recover the pseudo-average values, $\frac{\beta}{1-\beta}\bar{\mathbf{S}}_i^{\prime j}$, theoretically. Nevertheless, the real pseudo-average values taking the rounding effect into account are $(\frac{bs}{a(a-bs)} + \frac{1-a}{a}) \bar{\mathbf{S}}_i^{\prime j}$. To make the influence of rounding error to the minimum, the theoretical and practical pseudo average should be as close as possible. It leads to the optimization problem as
\begin{equation}
\underset{0\le \beta, \beta \neq 1}{\arg\min} \quad || f(\beta) \bar{\mathbf{S}}_i^{\prime j} - \frac{\beta}{1-\beta}\bar{\mathbf{S}}_i^{\prime j}|| = \underset{0\le \beta, \beta \neq 1}{\arg\min} \quad |f(\beta) - \frac{\beta}{1-\beta}|
\label{eq: minimum invariance optimization problem}
\end{equation}
where $f(\beta) = \frac{bs}{a(a-bs)} + \frac{1-a}{a}$. The optimization problem can be solved using the iterative method in \cref{eq: Jacobi Iteration} with a given initial guess for $\beta$. The suggested parameter, $\beta$, is as close to 1 as possible. In this situation, the term $\frac{1-a}{a}$ in $f(\beta)$ is quite small, so it is negligible compared to the dominant term. 

\newpage
\section{Optimal Accuracy Condition to Determine the Hyper-parameter $\beta$.}
\label{APP: Code for Optimal Accuracy Condition}

\definecolor{mygreen}{rgb}{0,0.6,0}
\definecolor{mygray}{rgb}{0.5,0.5,0.5}
\definecolor{mymauve}{rgb}{0.58,0,0.82}

To solve the optimal parameter in the nonlinear equation \ref{eq: nonlinear function}, the command \pyth{python optimal_para.py} is run for the below presented \textit{Python} code.

\lstset{ %
  backgroundcolor=\color{white},   
  basicstyle=\footnotesize,        
  breakatwhitespace=false,         
  breaklines=true,                 
  captionpos=bl,                    
  commentstyle=\color{mygreen},    
  deletekeywords={...},            
  escapeinside={\%*}{*)},          
  extendedchars=true,              
  frame=single,                    
  keepspaces=true,                 
  keywordstyle=\color{blue},       
  morekeywords={*,...},            
  numbers=left,                    
  numbersep=5pt,                   
  numberstyle=\tiny\color{mygray}, 
  rulecolor=\color{black},         
  showspaces=false,                
  showstringspaces=false,          
  showtabs=false,                  
  stepnumber=1,                    
  stringstyle=\color{orange},     
  tabsize=2,                       
}
\begin{lstlisting}[language=Python]
# optimal_para.py
import numpy as np
import torch
# To obtain the optimal alpha for PASA algorithm using fixed-point iteration.
# The Nonlinear Equation: beta / (1-beta) = f(beta), f(beta) = b*n / (a*(a-b*n))
def obtainInvPam(beta0, N, tp = torch.float16, cp = torch.float64):
    M0 = torch.tensor(1.0, dtype = cp) - beta0.type(cp) / N
    M1 = -beta0.type(cp) / N
    M0 = M0.type(tp)
    M1 = M1.type(tp)
    b = -M1.type(cp)
    a = M0.type(cp) + b
    Inv_Pam = b * N / (a * (a - b * N)) + (1 - a) / a
    return Inv_Pam
def optimal_beta(beta0, N, tol = 1.0e-8, tp = torch.float16, cp = torch.float64):
    err = 1.0
    iter = 0
    Inv_Pam = obtainInvPam(beta0, N, tp, cp)
    while (err > tol):
        Inv_Pam = obtainInvPam(beta0, N, tp, cp)
        beta = Inv_Pam / (1.0 + Inv_Pam)
        err = torch.abs(beta - beta0) / torch.abs(beta0)
        beta0 = beta * 1.0
        iter += 1
    return beta
if __name__ == "__main__":
    # float16
    print("=======================float16(1,5,10)==================")
    print("Initial beta = 1-1/2**4, 1-1/2**5, 1-1/2**6")
    bits = int(3)
    beta0 = torch.zeros(bits)
    for i in range(bits):
        beta0[i] = 1.0 - 1.0 / 2**(i+4)
    N = int(128)
    M = beta0.size()
    M = M[0]
    beta = torch.zeros(M)
    for i in range(M):
        beta[i] = optimal_beta(beta0[i], N, tp = torch.float16)
    print("========================Results:=========================")
    print(f"for float16, initial beta: {beta0}")
    print(f"for float16, beta: {beta}")
\end{lstlisting}

\newpage
\section{Numerical Accuracy Results for Random Benchmark Cases}
\label{APP: Numerical Results for Random Cases}

\begin{figure}[h]
    \centering
    \subfigure[Fixed Amplitude $Am = 0.5$]{
        \includegraphics[width=0.45\linewidth]{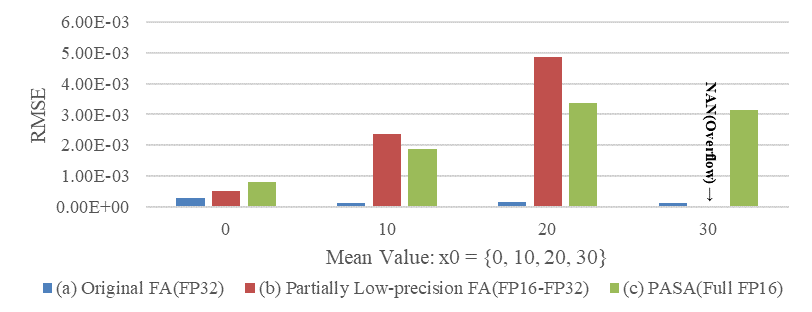}
    }
    \subfigure[Fixed Mean Value $x_0 = 20$]{
        \includegraphics[width=0.45\linewidth]{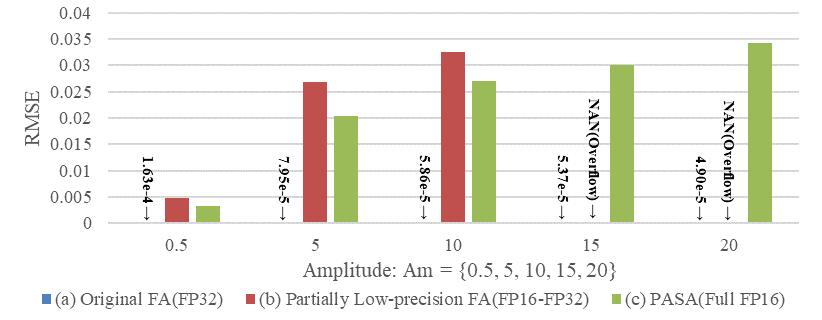}
    }
    \caption{RMSE Comparison for Three Precision Allocations of FA Algorithms for Uniform Random Data Distribution(Left Figure: Fixed Amplitude $Am$ and Varying Mean Value $x_0$; Right Figure: Fixed Mean Value $x_0$ and Varying Amplitude $Am$. The \textit{NAN} is not explicitly plotted and replaced by a text mark).}
    \label{fig:UniformDistri_RMSE}
\end{figure}

\begin{figure}[h]
    \centering
    \subfigure[Fixed Amplitude $Am = 10$]{
        \includegraphics[width=0.45\linewidth]{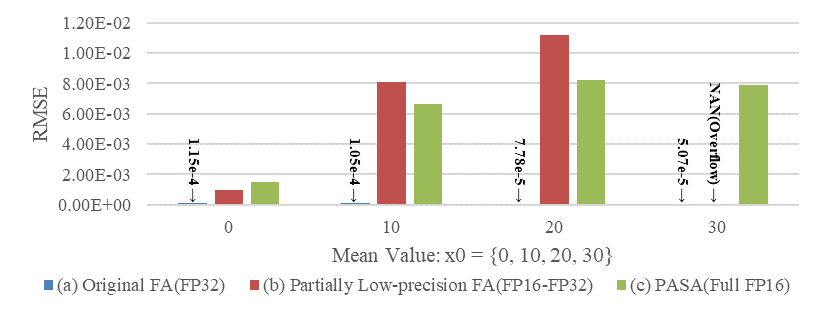}
    }
    \subfigure[Fixed Mean Value $x_0 = 20$]{
        \includegraphics[width=0.45\linewidth]{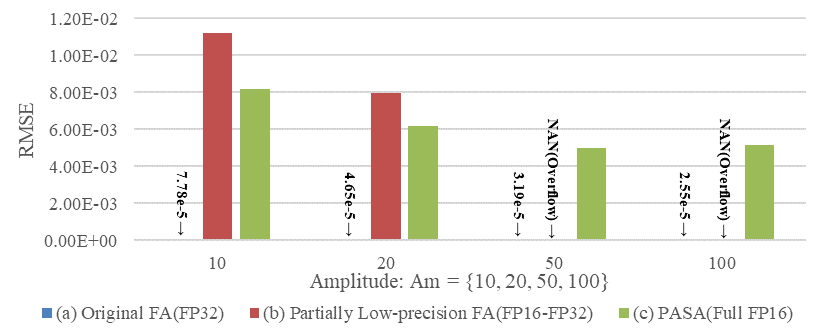}
    }
    \caption{RMSE Comparison for Three Precision Allocations of FA Algorithms for Hybrid Random Data Distribution(Left Figure: Fixed Amplitude $Am$ and Varying Mean Value $x_0$; Right Figure: Fixed Mean Value $x_0$ and Varying Amplitude $Am$. The \textit{NAN} is not explicitly plotted and replaced by a text mark).}
    \label{fig:HybridDistri_RMSE}
\end{figure}

The uniform and hybrid data distribution can be found in Equations \ref{eq: generation formulation for uniform random} and \ref{eq: generation formulation for hybrid random}. It is worth noting that the hybrid data distribution with the amplitude $Am=10$ and the mean value $x_0=0$ in Figure \ref{fig:HybridDistri_RMSE}a is same as the random data in FA3.0\autocite{shah2024flashattention}. As depicted in Figure \ref{fig:UniformDistri_RMSE}a, we choose different mean values for a fixed amplitude of $Am = 0.5$ for uniform random data distribution. The results indicate that overflow phenomenon appears when the mean value $x_0$ reaches $30$ only for Partially low-precision allocation of FA(FA16-FP32). In contrast, PASA has a similar capability of avoiding overflow with original high-precision allocation of FA(FP32) widely adopted in FA on GPU\autocite{dao2022flashattention,dao2023flashattention2,shah2024flashattention}. When it comes to the numerical accuracy for cases without overflow, it is observed that PASA has a smaller RMSE than Partially low-precision FA(FA16-FP32) for all cases with non-zero mean values, but larger than the original FA(FP32). Similar trends are also illustrated in Figure \cref{fig:UniformDistri_RMSE}b with increasing the amplitude $Am$ for a relatively small mean value $x_0 = 20$. Overflow occurs when $Am$ is larger than $10$ for Partially low-precision allocation of FA(FA16-FP32), while the phenomenon does not appear in other two precision allocation methods. For all cases in Figure \cref{fig:UniformDistri_RMSE}b, the numerical accuracy in PASA is higher than Partially low-precision allocation of FA(FA16-FP32), but is not as accurate as original high-precision allocation of FA(FP32).

 For random hybrid normal-Bernoulli distribution, when the amplitude $Am$ is fixed to a small value $10$, the trends for the overflow and numerical accuracy in Figure \ref{fig:HybridDistri_RMSE}a are consistent with these in Figure \cref{fig:UniformDistri_RMSE}a for uniform distribution. When we fix the mean value $x_0$ to 20, the overflow appears for the amplitude $Am$ exceeding to $20$ for partially low-precision allocation of FA(FP16-FP32) in Figure \ref{fig:HybridDistri_RMSE}b, while PASA and original high-precision allocation of FA(FP32) still keep normal computation free from overflow or final \textit{NAN} results. 

\newpage
\section{Statistics for the \textit{NAN} Appearance Percentages of FA Output for Random Benchmark Datasets in Partially Low-precision FA(FP16-FP32).}
\label{APP: Statistics for the NAN Percentages}

\begin{table}[ht]
\caption{\textit{NAN} Percentages of FA Output for the Datasets with Uniform and Hybrid Random Distributions for Partially Low-precision FA(FP16-FP32)(Uniform: Uniform Random Distribution in Equation \ref{eq: generation formulation for uniform random}; Hybrid: Hybrid Normal-Bernoulli Random Distribution in Equation \ref{eq: generation formulation for hybrid random}).}
\label{Tab: Statistics for the NAN Percentages-high-per FA}
\vskip 0.15in
\begin{center}
\begin{small}
\begin{sc}
\begin{tabular}{cccccc}
\toprule
$N0$   & Distribution Type & Mean Value, $x_0$ & Amplitude, $Am$ & \textit{NAN} Percentage & Overflow$?$ \\
\midrule
$1$  &   Uniform   &   $30$ & $0.5$   &  $100\%$  & Yes \\
$2$  &   Uniform   &   $20$ & $15$    &  $0.12\%$ & Yes \\
$3$  &   Uniform   &   $20$ & $20$    &  $8.14\%$ & Yes  \\
$4$  &   Hybrid    &   $30$ & $10$    &  $100\%$  & Yes  \\
$5$  &   Hybrid    &   $20$ & $50$    &  $0.04\%$ & Yes   \\
$6$  &   Hybrid    &   $20$ & $100$   &  $1.11\%$ & Yes   \\
\bottomrule
\end{tabular}
\end{sc}
\end{small}
\end{center}
\vskip -0.1in
\end{table}

\newpage
\onecolumn
\section{Cloud Maps for Real LMs(Qwen2 and stable-video-diffusio-IMG2VID Models).}
\label{APP: Cloud Maps for Real LMs}

\begin{figure}[ht]
    \centering
    \subfigure[Query Matrix, $\mathbf{Q}$]{
        \includegraphics[width=0.3\linewidth]{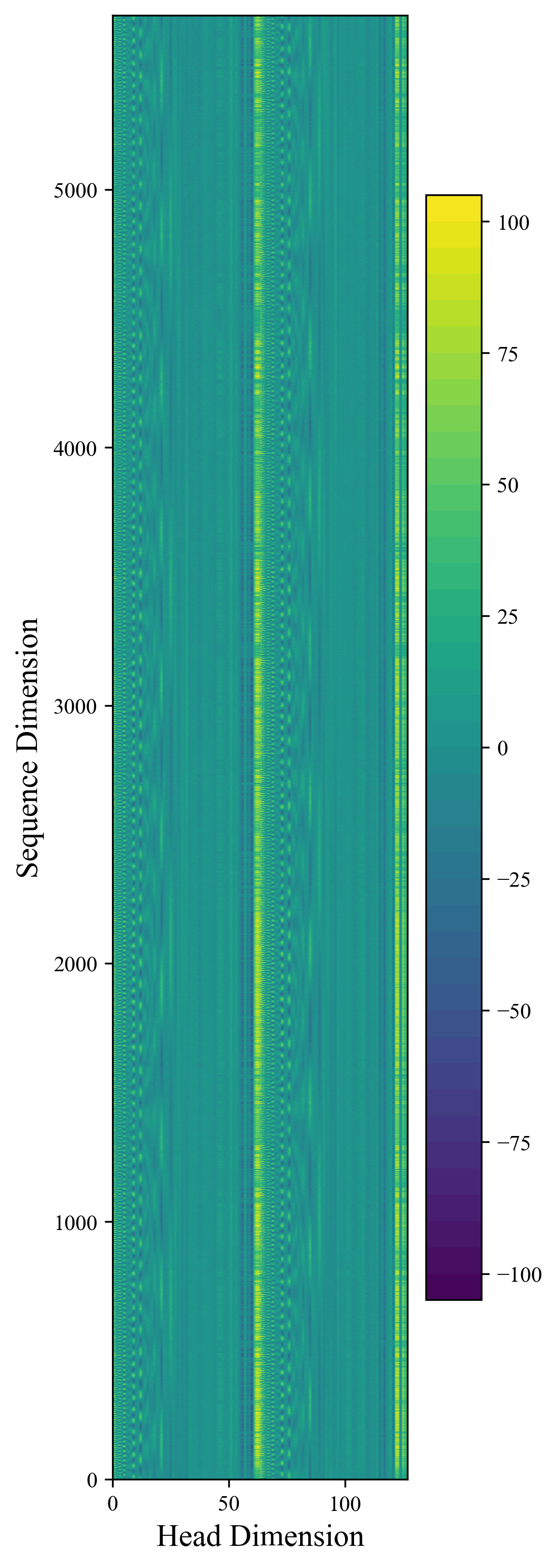}
    }
    \subfigure[Key Matrix, $\mathbf{K}$]{
        \includegraphics[width=0.3\linewidth]{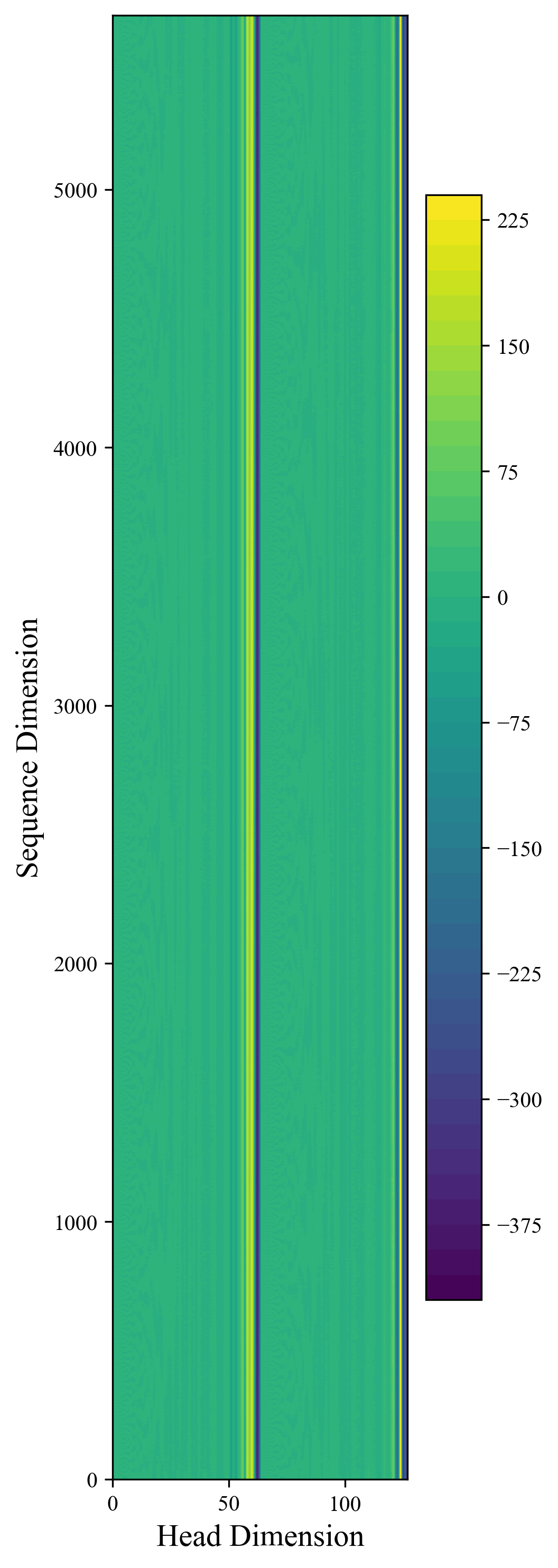}
    }
    \subfigure[Pseudo-average Shifted Key, $\mathbf{K}^T\mathbf{M}$]{
        \includegraphics[width=0.3\linewidth]{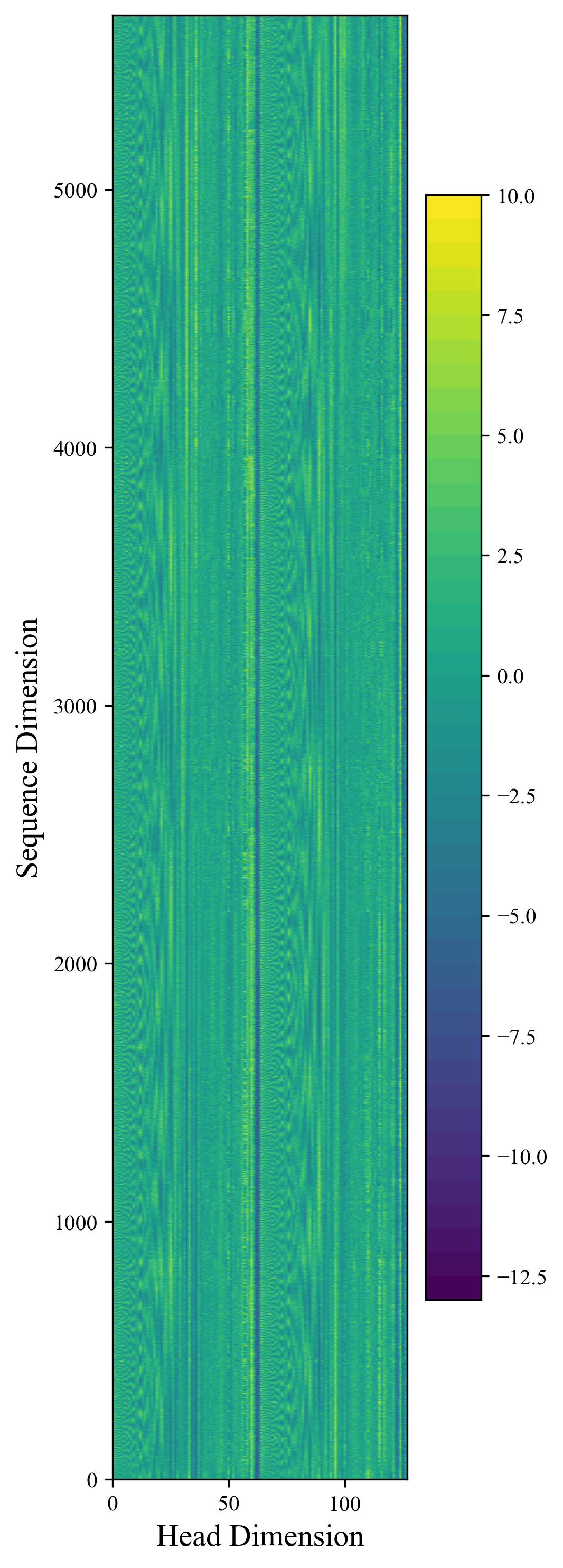}
    }
    \caption{The Data Distribution of Query and Key Matrices for Qwen2 Overflow Cases($Batch=0$, $Head=10$, hyper-parameter - $\beta = 0.984497$).}
    \label{fig:cloud pictureS for The original QK and preprocessed K in Qwen2-7B}
\end{figure}

\begin{figure}[ht]
    \centering
    \subfigure[Query Matrix, $\mathbf{Q}$]{
        \includegraphics[width=0.3\linewidth]{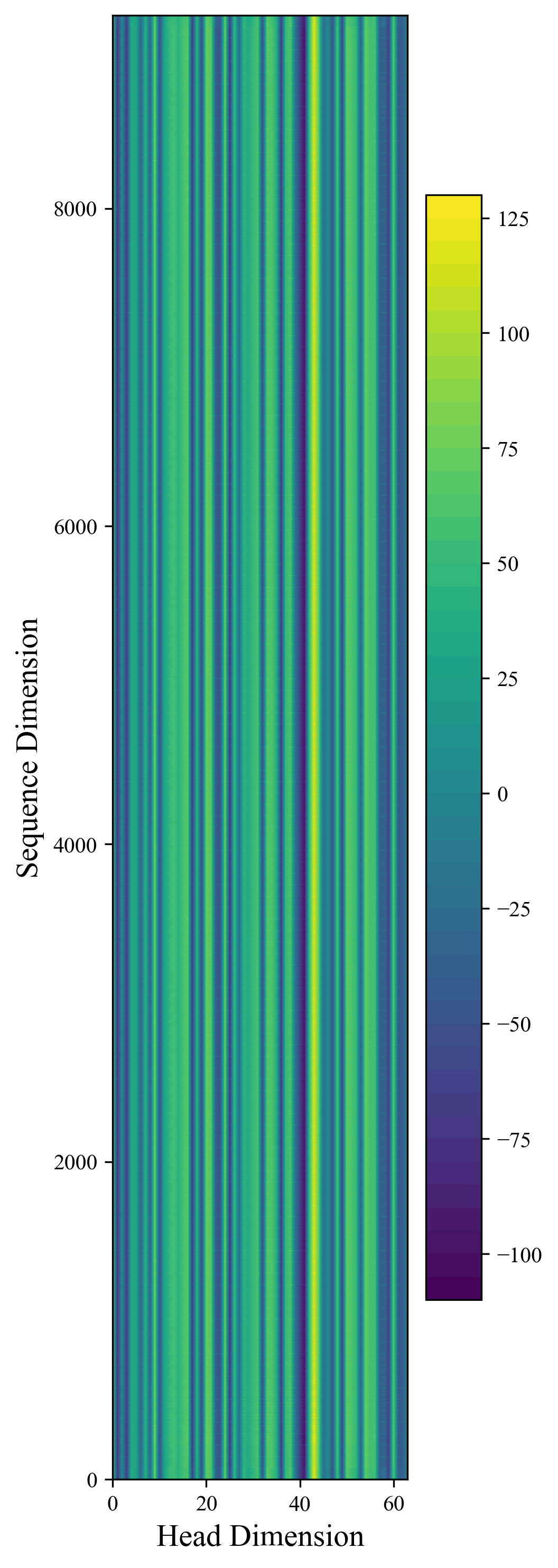}
    }
    \subfigure[Key Matrix, $\mathbf{K}$]{
        \includegraphics[width=0.3\linewidth]{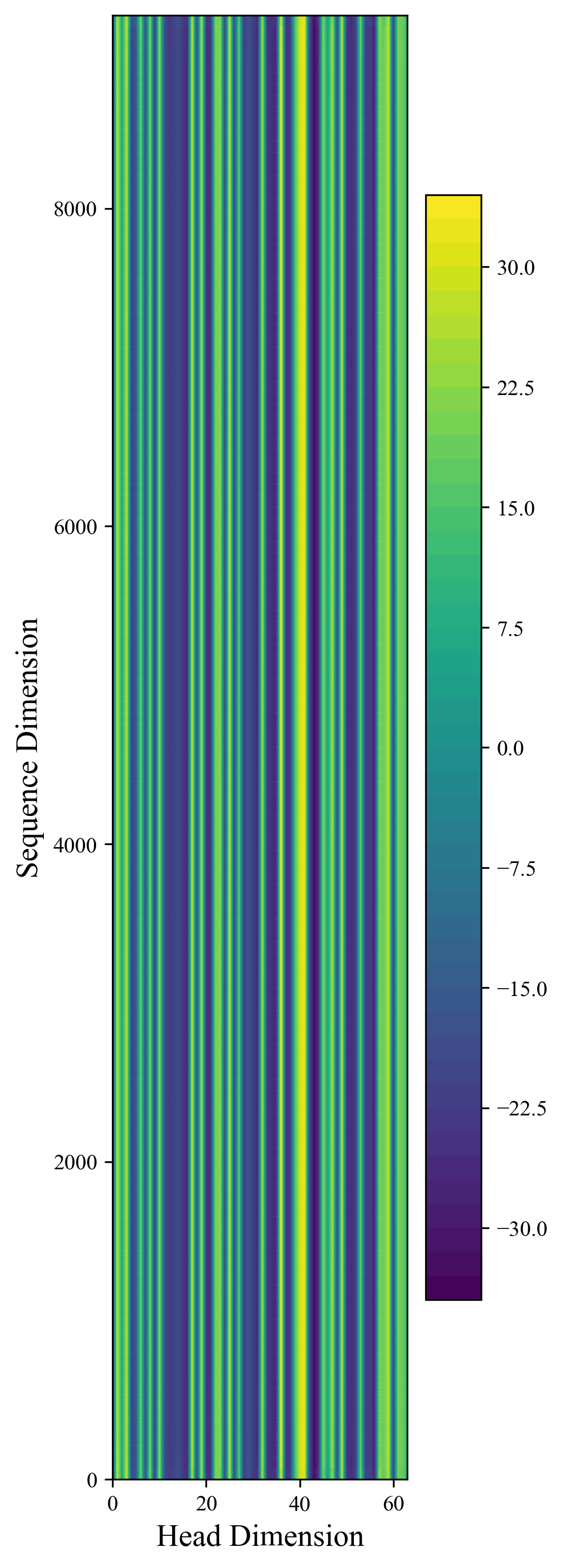}
    }
    \subfigure[Pseudo-average Shifted Key, $\mathbf{K}^T\mathbf{M}$]{
        \includegraphics[width=0.3\linewidth]{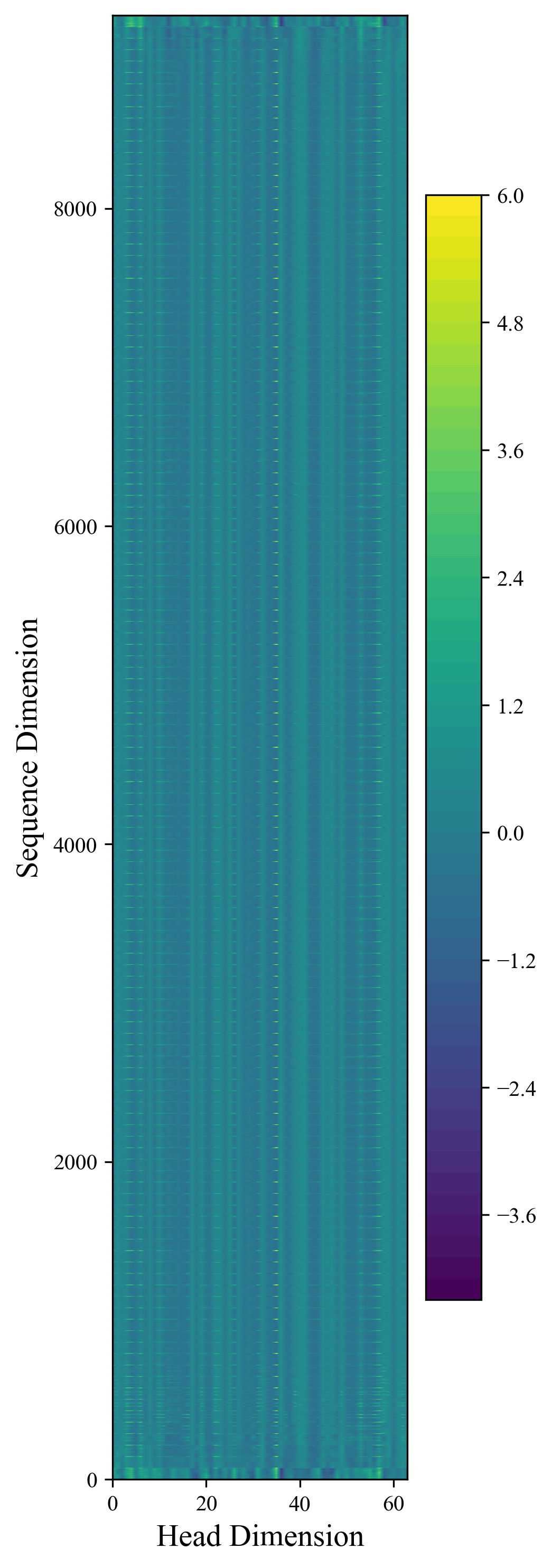}
    }
    \caption{The Data Distribution of Query and Key Matrices for SVD-IMG2VID Overflow Cases($Batch=1$, $Head=4$, hyper-parameter - $\beta = 0.984497$).}
    \label{fig:cloud pictureS for The original QK and preprocessed K in IMG2VID}
\end{figure}

\begin{figure}[ht]
    \centering
    \subfigure[Attention Score Matrix: $\mathbf{S}=\mathbf{QK}^T$]{
        \includegraphics[width=0.48\linewidth]{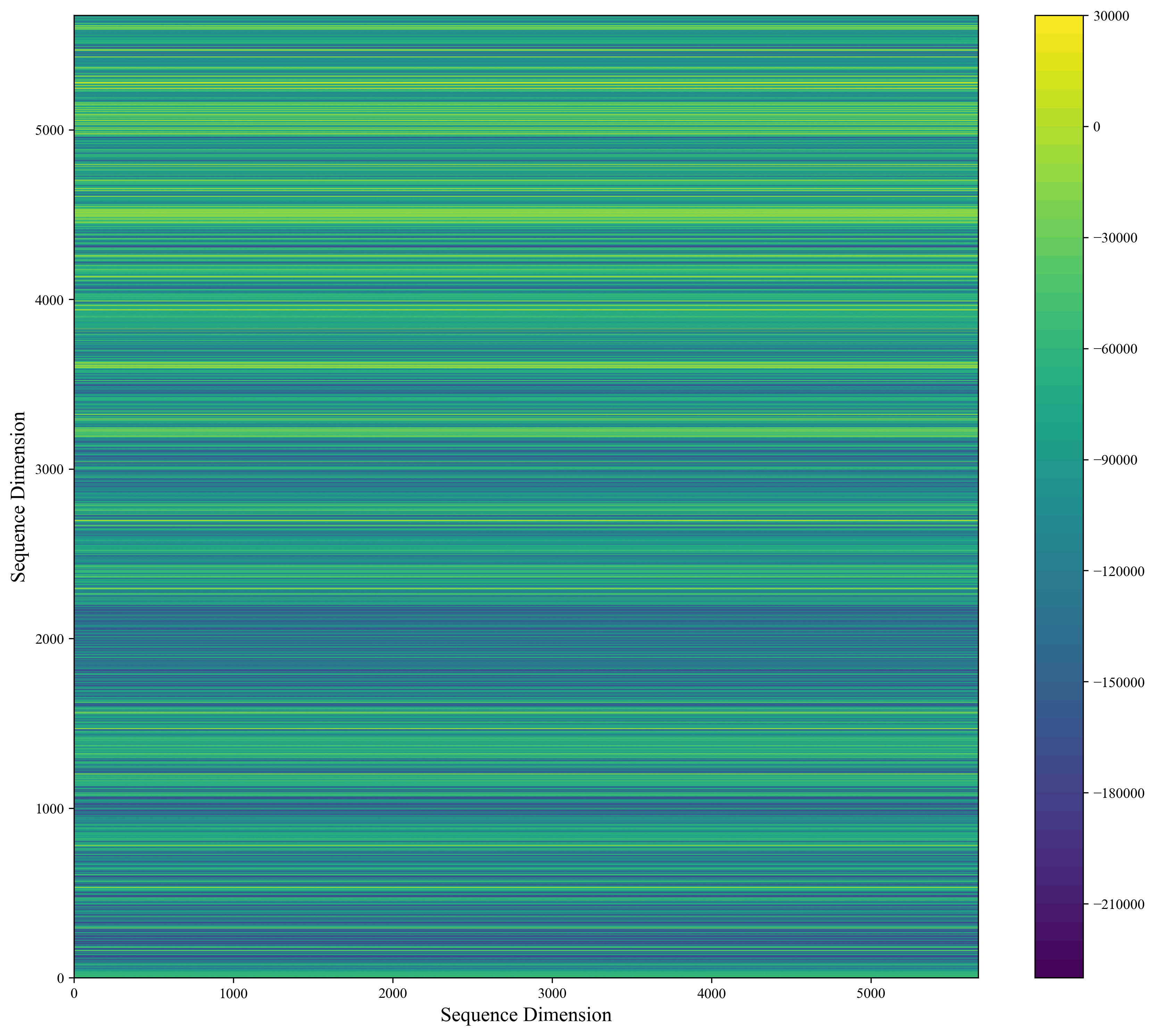}
    }
    \subfigure[Pseudo-average Shifted Attention Score Matrix: $\mathbf{SM}$]{
        \includegraphics[width=0.48\linewidth]{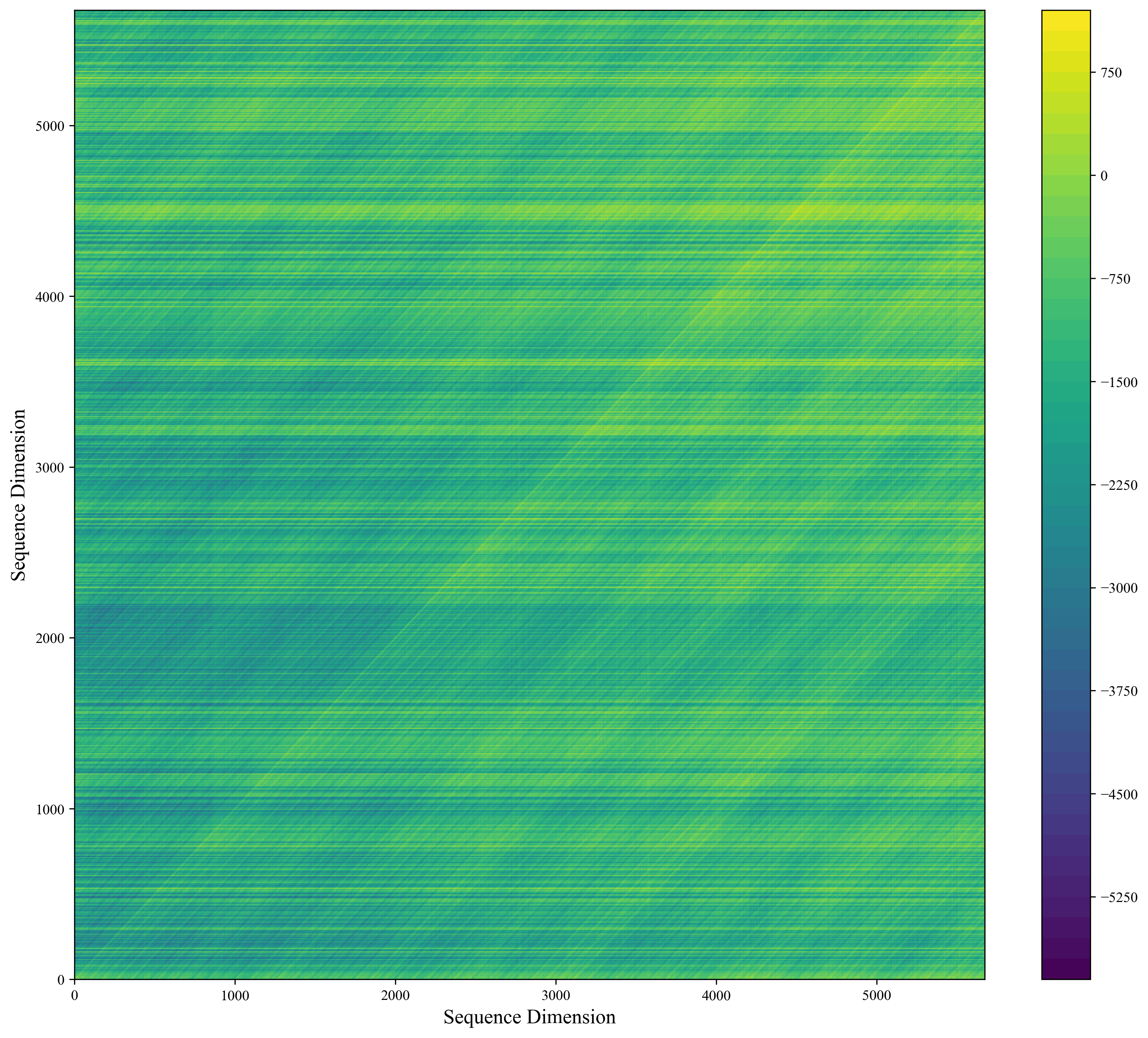}
    }
    \caption{The Data Distribution of Original and Preprocessed Attention score matrices for Qwen2 Overflow Cases($Batch=0$, $Head=10$, hyper-parameter - $\beta = 0.984497$).}
    \label{fig:cloud picture for the attention score matrix S for Qwen2-7B}
\end{figure}

\begin{figure}[h]
    \centering
    \subfigure[Attention Score Matrix: $\mathbf{S}=\mathbf{QK}^T$]{
        \includegraphics[width=0.48\linewidth]{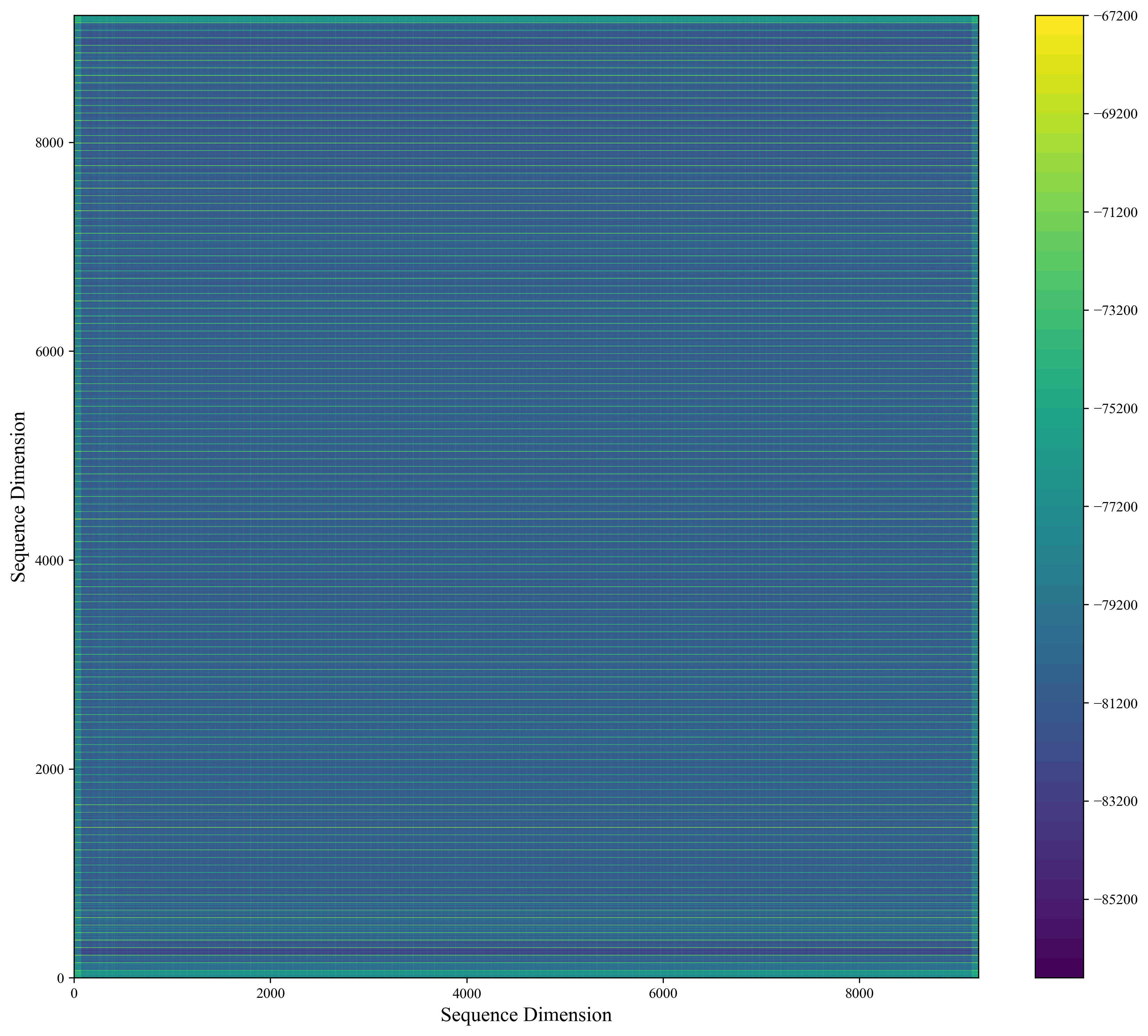}
    }
    \subfigure[Pseudo-average Shifted Attention Score Matrix: $\mathbf{SM}$]{
        \includegraphics[width=0.48\linewidth]{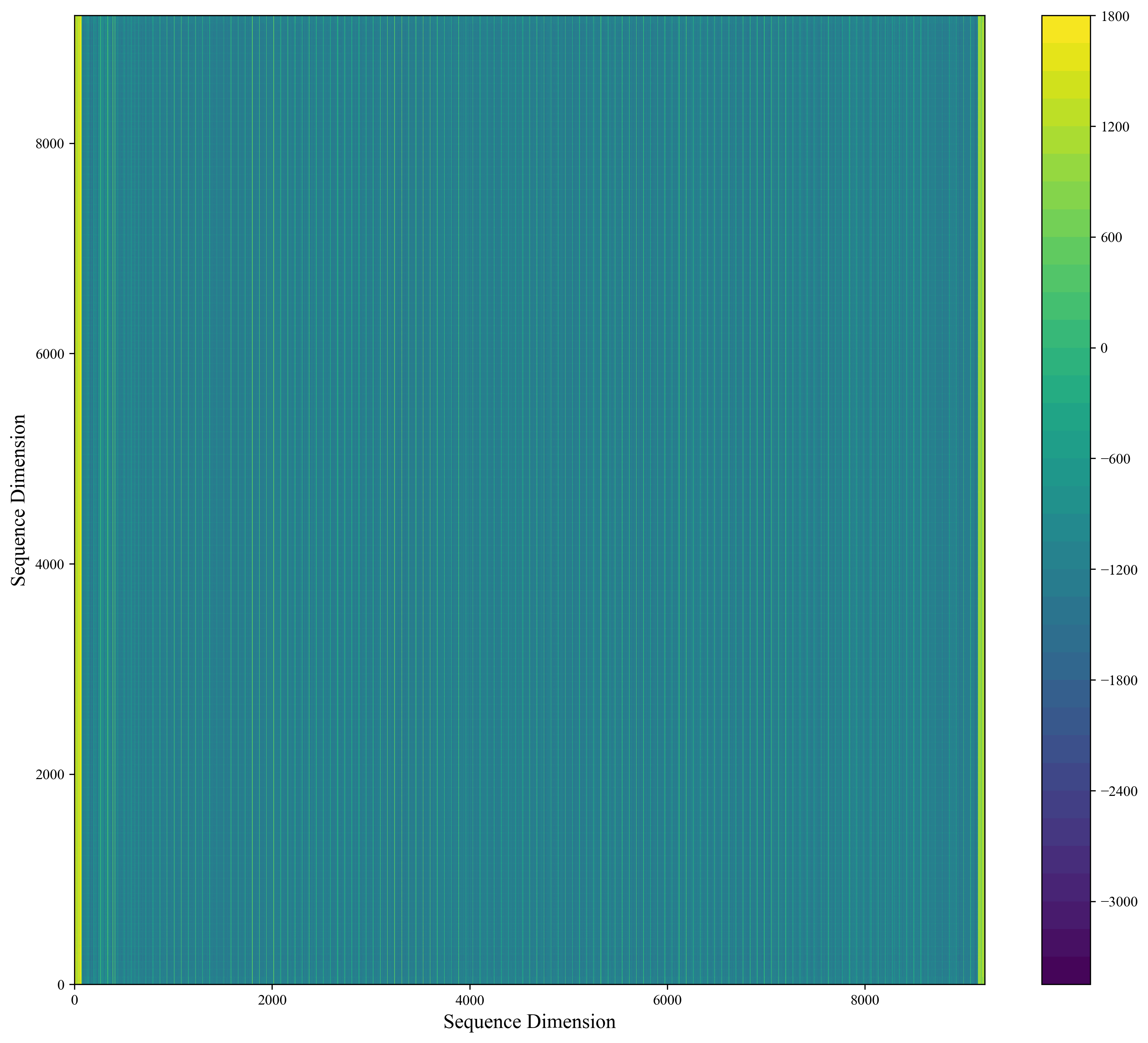}
    }
    \caption{The Data Distribution of Original and Preprocessed Attention score matrices for SVD-IMG2VID Overflow Cases($Batch=1$, $Head=4$, hyper-parameter - $\beta = 0.984497$).}
    \label{fig:cloud picture for the attention score matrix S for IMG2VID}
\end{figure}

\newpage
\onecolumn
\section{Experimental Parameter Setup for the LM Inference Cases(Qwen2-7B and SVD-IMG2VID).}
\label{APP: Prompt Information for LM Inference Cases}

The prompt from the LongBench\autocite{bai2023longbench} dataset is applied to validate the inference accuracy with PASA for Qwen2-7B language model. It is found that almost all the test cases from the LongBench\autocite{bai2023longbench} will lead to the occurrence of overflow for partially low-precision or naively full precision allocation of FA. We only present the representative one with more than \textit{5k} sequence length. The simplified prompt information is:

\textit{\textbf{[Prompt]}: Answer the question based on the given passage. Only give me the answer and do not output any other words. The following are some examples. \{Passage: Adam's apple The laryngeal prominence (commonly referred to as Adam's apple), a feature of the human} \textit{... ...} \textit{The visitor center is open daily (except Thanksgiving Day, December 25, and January 1) from 9:00 a.m. to 5:00 p.m, with extended hours between Memorial Day and September 30. During the summer season the visitor center is open until the laser light show, One River, Many Voices, ends. Show times vary, learn more $>>$ Question: In which country is the Grand Coulee Dam Answer:\}}

Both the original output from Qwen2-7B inference with high-precision attention and the output with PASA in FP16 precision are: \textit{United States}. It indicates that the PASA has a similar inference accuracy with the original high-precision attention for language model.

When it comes to the SVD-IMG2VID test cases, the benchmark input is given in the picture form. The official python script in \href{https://huggingface.co/docs/diffusers/using-diffusers/svd}{Huggingface} for the inference is given as:
\begin{lstlisting}[language=Python]
import torch
from diffusers import StableVideoDiffusionPipeline
from diffusers.utils import load_image, export_to_video

pipeline = StableVideoDiffusionPipeline.from_pretrained(
    "stabilityai/stable-video-diffusion-img2vid", torch_dtype=torch.float16, variant="fp16"
)
pipeline.enable_model_cpu_offload()
pipeline.unet.enable_forward_chunking()

image = load_image("https://huggingface.co/datasets/huggingface/documentation-images/resolve/main/diffusers/svd/rocket.png")
image = image.resize((1024, 576))

generator = torch.manual_seed(42)
frames = pipeline(image, decode_chunk_size=1024, generator=generator, num_frames=25).frames[0]

export_to_video(frames, "video.mp4", fps=7)
\end{lstlisting}

\end{document}